\documentclass[10pt,twocolumn,letterpaper]{article}
\usepackage[pagenumbers]{cvpr}
\usepackage{colortbl} %
\usepackage{multirow} %

\usepackage{cuted} %

\usepackage{currfile} %

\usepackage{caption} %

\usepackage{algorithm}
\usepackage[noend]{algpseudocode} %
\usepackage[most]{tcolorbox} %

\definecolor{cvprblue}{rgb}{0.21,0.49,0.74}
\usepackage[breaklinks,colorlinks,allcolors=cvprblue]{hyperref}
\hypersetup{
  pdftitle={RoboHarness-RL: Agentic Reinforcement Learning for VLM-Based Robot Manipulation},
  pdfauthor={Liyang Li, Mingyu Liu, Xiaoman Li, Zhiyue Zhao, Jin-Chuan Shi, Canyu Zhao, Haokai Zhang, Jinyuan Tian, Linhao Zhong, Hao Chen, Chunhua Shen},
  pdfsubject={Agentic reinforcement learning for VLM-based robot manipulation}
}
\title{RoboHarness-RL: Agentic Reinforcement Learning for VLM-Based Robot Manipulation}
\author{
  \normalsize\bfseries Liyang Li\textsuperscript{*}\quad Mingyu Liu\textsuperscript{*}\quad Xiaoman Li\quad Zhiyue Zhao\quad Jin-Chuan Shi\quad Canyu Zhao\\[3pt]
  \normalsize\bfseries Haokai Zhang\quad Jinyuan Tian\quad Linhao Zhong\quad Hao Chen\textsuperscript{$\dagger$}\quad Chunhua Shen\textsuperscript{$\dagger$}\\[7pt]
  \normalsize\normalfont Zhejiang University, State Key Lab of CAD \& CG
}
\begin{document}
\twocolumn[{%
\renewcommand\twocolumn[1][]{#1}%
\maketitle
\begin{center}
    \vspace*{-12pt}
    \centering
    \captionsetup{type=figure}
    \includegraphics[width=\textwidth]{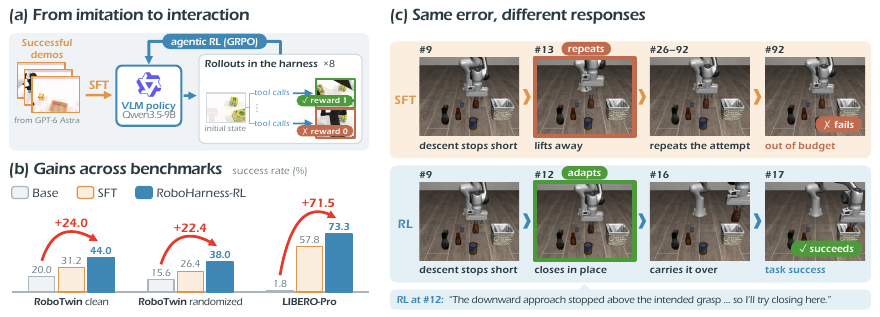}
    \captionof{figure}{RoboHarness-RL trains VLMs to act through tool calls, without a VLA policy. \textbf{(a)}~SFT on teacher demonstrations, then agentic RL. \textbf{(b)}~Red arrows: gain over the base model. \textbf{(c)}~After the same error, SFT repeats it and fails; RL adapts and succeeds.}
    \label{fig:teaser}
\end{center}%
\vspace{0.5em}
}]

{\renewcommand{\thefootnote}{\fnsymbol{footnote}}\footnotetext[1]{Equal contribution. \textsuperscript{$\dagger$}Corresponding authors.}}
\begin{abstract}
Agentic harnesses enable frontier vision-language models (VLMs) to manipulate robots through perception, planning, and control tools, yet achieving comparable performance with compact open-weight VLMs remains challenging.
We present \textbf{RoboHarness-RL}, an agentic reinforcement learning framework for improving VLM-based robot manipulation through tool calls.
Starting from supervised fine-tuning on successful teacher demonstrations, the VLM further improves through multi-turn reinforcement learning with task-outcome rewards, without relying on vision-language-action models or world action models.
To support efficient multi-turn training, we develop \textbf{LiteHarness}, which uses hierarchical memory to keep interaction contexts compact while preserving task performance.
With Qwen3.5-9B, RL raises success from 31.2\% to 44.0\% on RoboTwin 2.0 and from 57.8\% to 73.3\% on LIBERO-Pro; these gains extend to randomized RoboTwin scenes unseen during training (26.4\% to 38.0\%).
Behavioral analysis further shows that RL reduces repetitive-call rates by up to $4\times$, shortens successful executions, and improves recovery from negative feedback.
Project page: \url{https://aim-uofa.github.io/RoboHarness-RL/}.

\end{abstract}

\section{Introduction}
\label{sec:intro}

Agentic harnesses let vision-language models (VLMs) manipulate robots through tool calls for perception, planning, and control~\cite{li2026robo,ora2026code,chen2026show}.
Robot systems have traditionally separated a slow model that reasons about the task (System~2) from a fast learned policy that produces motor actions (System~1). 
In a harness, this boundary blurs: the VLM not only decides what to do next, but also specifies end-effector poses and gripper commands, so its decisions directly shape how the robot moves.

Yet achieving comparable performance with compact open-weight VLMs remains challenging.
On randomized RoboTwin 2.0 scenes, the ORA harness reaches 69.2\% success with DeepSeek-V4.1-Flash but only 6.8\% with the open-weight Qwen3.5-9B (Sec.~\ref{sec:comparison}).
Approaches to improving compact VLM agents include refining the harness~\cite{wang2026harnesspai,gong2026tinyastral} or distilling frontier models through supervised learning~\cite{li2026robo,chen2026show}.
Building on these advances, we investigate whether reinforcement learning from the model's own interactions can further improve manipulation performance beyond supervised imitation.

\begin{figure}[!t]
    \centering
    \includegraphics[width=\linewidth]{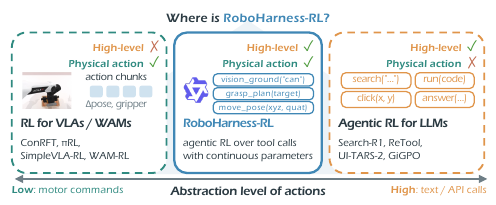}
    \caption{Where is RoboHarness-RL? It optimizes robot tool calls, which are higher-level than the motor commands of VLAs and finer-grained than the text actions of LLM agents.}
    \label{fig:positioning}
\end{figure}

Reinforcement learning (RL) has improved vision-language-action models (VLAs)~\cite{chen2025conrft,wagenmaker2025steering,li2025gr} and world action models (WAMs)~\cite{qian2026wam,ye2026halo} beyond imitation.
In parallel, agentic RL trains language models through multi-turn tool use in digital environments such as web search~\cite{jin2025search} and graphical interfaces~\cite{wang2025ui}.
Harness-based manipulation lies between the two (Fig.~\ref{fig:positioning}): the VLM uses high-level tool calls as in agentic RL, while controlling physical robot actions.
Concurrent approaches include RL with a learned VLA executor~\cite{liu2026agenticroborl} and single-turn RL for control-program generation~\cite{fu2026cap}.
We instead train the VLM to specify robot actions itself through multi-turn tool calls.

Two challenges make RL costly in this setting.
First, every model call carries a growing interaction history, including camera images, across episodes of up to 200 calls, increasing the cost of both rollout collection and policy updates.
Second, tasks that the current policy consistently solves or fails may provide little group-relative learning signal, yet still consume rollout budget.

We address both with \textbf{RoboHarness-RL}, an agentic RL framework that trains VLMs to manipulate through tool calls, with no VLA or WAM generating actions.
After fine-tuning on teacher demonstrations, the VLM learns through multi-turn GRPO with task-outcome rewards.
Interaction runs through \textbf{LiteHarness}, whose hierarchical memory keeps recent interactions in detail, summarizes older actions, and records failures separately.
Compared with compressed full history, it roughly halves the prompt tokens per call while achieving higher success.
Adaptive scheduling postpones solved tasks, removes persistently failed ones, and allocates more rollouts to tasks with a learning signal.

With Qwen3.5-9B~\cite{qwen2026qwen35}, RL raises success from 31.2\% after SFT to \textbf{44.0\%} on RoboTwin 2.0~\cite{chen2025robotwin} and from 57.8\% to \textbf{73.3\%} on LIBERO-Pro.
Although trained only in clean scenes, the policy also improves from 26.4\% to \textbf{38.0\%} in randomized RoboTwin scenes.
RL also succeeds on tasks that SFT fails to solve in evaluation.
Behavioral analysis shows how RL changes the way the model uses the harness: on RoboTwin, repetitive-call rates drop from about 10\% to under 3\%, jointly successful episodes require about a third fewer model calls, and the policy more often recovers to complete the task after execution failures (Fig.~\ref{fig:teaser}c).

Our main contributions are as follows:
\begin{itemize}
\item We propose \textbf{RoboHarness-RL}, an agentic RL framework that trains VLMs to manipulate robots through multi-turn tool calls without relying on VLAs or WAMs.
\item We design \textbf{LiteHarness} and adaptive task scheduling for efficient multi-turn RL. Hierarchical memory roughly halves prompt tokens per call and improves success.
\item We show that RL improves over supervised imitation by \textbf{12.8 and 15.5 percentage points} on RoboTwin 2.0 and LIBERO-Pro, respectively. These gains extend to randomized scenes unseen during training. RL also reduces repetitive tool calls, shortens successful executions, and improves recovery after negative feedback.
\end{itemize}

\section{Related Work}
\label{sec:related}

\begin{figure*}[t]
    \centering
    \includegraphics[width=\textwidth]{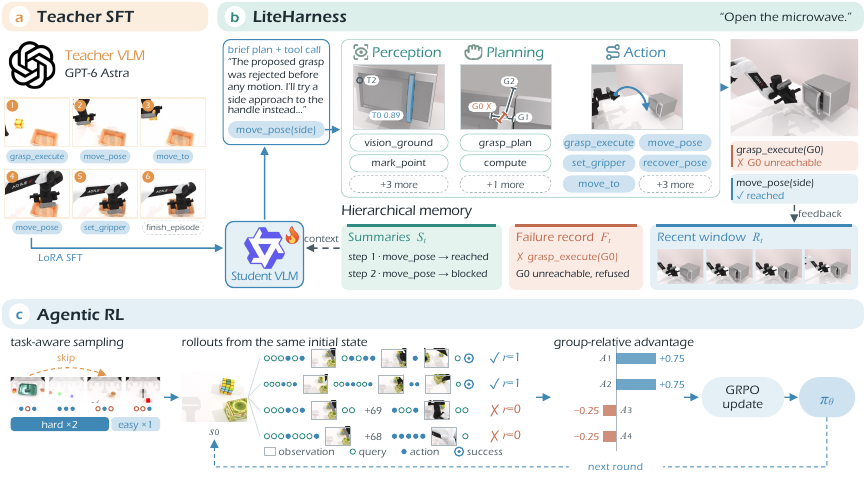}
    \caption{Overview of RoboHarness-RL. \textbf{(a)}~SFT on successful teacher episodes initializes the student VLM. \textbf{(b)}~In LiteHarness, the student issues tool calls to fixed perception, planning, and action tools; feedback is kept in a hierarchical memory that forms the next context. \textbf{(c)}~Agentic RL samples groups of rollouts with per-task scheduling and updates the policy with GRPO.}
    \label{fig:framework}
\end{figure*}

\subsection{Agentic harnesses for robot manipulation}
\label{sec:rw-harness}
Early language-guided systems compose predefined skills~\cite{ahn2022can,huang2022inner,singh2022progprompt}, generate policy code~\cite{liang2023code}, or guide motion planning with spatial representations~\cite{huang2023voxposer}; hierarchical VLM planning also supports long-horizon mobile manipulation~\cite{wang2026odyssey}.

Recent harnesses enable closed-loop control through code and tools.
Code-based systems compose perception and control primitives and revise programs from execution feedback~\cite{fu2026cap,naouali2026vlcp,jia2026agent,gao2026hexashelluniversalharnessrobot}.
Tool-based systems expose perception, planning, and control functions for direct invocation~\cite{li2026robo,ora2026code,chen2026show,zhang2026world,yuan2026democratizing,zhang2026unexpected}.
Other harnesses connect VLM motion commands to deterministic control with asynchronous monitoring~\cite{li2026motormind}.

Across episodes, agents refine code through repeated trials~\cite{xiao2026enpire,wen2026roborsi}, evolve harnesses around frozen models~\cite{wang2026harnesspai,gong2026tinyastral}, or retain experience as persistent memory~\cite{wang2026regenharness,liu2026phyagentos}, transferable spatial lessons~\cite{zhang2026spatialmemory}, and reusable skills~\cite{huang2026scaling}.
RoboHarness-RL instead improves model decisions through multi-turn RL.

\subsection{Agentic reinforcement learning for language models}
\label{sec:rw-agentic-rl}
Agentic RL trains language models through multi-turn interaction, including search~\cite{jin2025search,zheng2025deepresearcher}, tool-assisted reasoning~\cite{feng2026retool,li2025torl}, and graphical or web interfaces~\cite{wang2025ui,wei2025webagent}.
Broader studies examine training stability, exploration, and rollout design~\cite{wang2025ragen,xi2025agentgym}, while fine-grained credit assignment combines trajectory-level advantages with step-level estimates~\cite{feng2026group}.

Training frameworks decouple agent execution from policy optimization through standardized interfaces or rollout services~\cite{luo2025agent,zhang2026prorl}.
They record model interactions for training while leaving tool use, context management, and execution to the harness~\cite{xu2026polar,he2026agent}.
We adopt this approach for robot manipulation.

\subsection{Training embodied agents}
\label{sec:rw-training}
Robot policies use sparse plans with action experts~\cite{liu2025gae}, sparse trajectory interfaces~\cite{huang2026notvla}, or joint-aligned actions and adversarial learning for cross-embodiment demonstration transfer~\cite{zhao2026advdex}.
Yet demonstration learning can leave modality shortcuts~\cite{liu2026perturbot} and insufficient recovery-state coverage~\cite{liu2026perfect}.
RL improves VLAs and diffusion policies beyond imitation~\cite{chen2025conrft,wagenmaker2025steering,li2025gr,lu2025vla,li2026simplevla,wang2026learning,intelligence2025pi}, and extends to world action models~\cite{qian2026wam,ye2026halo}.
SFT and multi-turn RL are also studied for embodied agents acting on visual feedback~\cite{li2026look}.

For robot harnesses, teacher distillation transfers frontier-model capabilities to compact agents~\cite{li2026robo,chen2026show,gong2026tinyastral}.
CaP-RL studies single-turn RL for control programs, while Guava explores RL refinement on two selected manipulation tasks~\cite{fu2026cap,liu2026guava}.
AgenticRoboRL~\cite{liu2026agenticroborl} invokes VLAs through RPent and generates code with privileged state through CaP-X; PyRUA-Lean Coder~\cite{si2026fewer} generates Python programs over primitives including frozen VLAs.
CaPtain combines SFT and multi-turn RL for code-as-policy agents with a privileged-state critic~\cite{fan2026captain}, while ZAP uses manually designed progress rewards~\cite{pikus2026pretrainingknowssimulationteaches}.
We use direct tool calls, task-success rewards, and general tool-use penalties, without VLAs, privileged-state critics, or task-specific reward shaping.
We train and evaluate on RoboTwin 2.0 and LIBERO-Pro with hierarchical memory and adaptive task scheduling.

\section{Method}
\label{sec:method}

\subsection{Overview}
\label{sec:method-overview}

RoboHarness-RL trains a VLM policy through multi-turn interaction with a robot environment.
Interaction is supported by \textbf{LiteHarness}, a lightweight harness built on the tools of ORA~\cite{ora2026code} and designed for efficient agentic RL.
Figure~\ref{fig:framework} illustrates the harness and the two training stages.

For a manipulation task with instruction $x$, a VLM policy $\pi_\theta$, parameterized by $\theta$, interacts with the environment through LiteHarness.
At model call $t$, the policy generates a response $y_t\sim\pi_\theta(\cdot\mid x,h_t)$, where $h_t$ denotes the interaction context provided by LiteHarness for that call.
Tool calls within these responses provide access to perception, planning, and control, with execution feedback guiding subsequent decisions.
The resulting interaction is represented by a trajectory $\tau=((h_1,y_1),\ldots,(h_T,y_T))$, where $T$ is the number of model calls until the episode ends.

Training proceeds in two stages: supervised fine-tuning on teacher demonstrations establishes basic manipulation procedures, followed by multi-turn agentic RL to improve decision-making through interaction.
Both stages optimize only the VLM policy, keeping the perception models and motion planners fixed.
Section~\ref{sec:lite-ora} describes LiteHarness, Section~\ref{sec:training} the two training stages, and Section~\ref{sec:task-aware-sampling} interaction budgets and task scheduling.

\subsection{LiteHarness: A Lightweight Agent Harness}
\label{sec:lite-ora}

LiteHarness is built around one constraint: every model call carries the interaction history, so the length of this context drives the cost of both rollout collection and policy optimization.
It keeps the context compact in two ways, through concise tool descriptions and through hierarchical memory.

\paragraph{Tool interface.}
The policy acts on the environment only through tools, which fall into two kinds.
Read-only queries return information without moving the robot, such as object masks from SAM~3~\cite{carion2026sam}, grasp candidates from Contact-GraspNet~\cite{sundermeyer2021contact}, or geometric measurements.
Robot-action tools end the policy's turn and execute grasp selections, end-effector targets, or gripper commands specified by the VLM.
Motion execution uses cuRobo~\cite{sundaralingam2023curobo} planning or local control, while gripper commands control opening and closing.
Concise descriptions of each tool's purpose and arguments keep the fixed context short.
Full interface specifications are provided in Appendix~\ref{sec:app-tools}.

\paragraph{Hierarchical memory.}
Hierarchical memory controls context growth by retaining different parts of the interaction history at different levels of detail.
Without it, feedback from every query and action would accumulate, and the policy often issues several queries before each robot action.
Omitting fixed instructions, the context is written as
\begin{equation}
    h_t = [S_t;\,F_t;\,R_t],
    \label{eq:hierarchical-memory}
\end{equation}
where $S_t$, $F_t$, and $R_t$ denote historical summaries, failure records, and recent interactions, respectively.
Here, $[\,\cdot\,;\,\cdot\,]$ denotes concatenation, and all memory components contain only information available before model call $t$.

$S_t$ summarizes actions older than the recent window and their outcomes to maintain a concise account of task progress.
To discourage repeated unsuccessful attempts, $F_t$ consolidates failed actions and rejected calls with their key arguments and reported causes, keeping prior failures accessible after their details leave the window.
Recent memory $R_t$ retains the current observation and up to $K$ preceding observations, together with their associated responses and tool feedback.
Earlier queries within this window are compressed while preserving key outputs, such as coordinates and candidate identifiers, for subsequent tool use.

Images use a shorter window: only the current and two preceding steps can contribute images.
Tool-feedback images from previous observations are discarded, and each request is capped at $B_{\mathrm{img}}$ images, removing the oldest first.
All memory operations are deterministic and require no additional model calls; retention and compression settings are provided in Appendix~\ref{sec:app-memory}.

\subsection{Training with Teacher Demonstrations and Agentic RL}
\label{sec:training}

\begin{table*}[t]
    \centering
    \caption{Success rates (\%) with LiteHarness; separate training per benchmark, clean-only on RoboTwin. RoboTwin: 250 episodes/condition; LIBERO-Pro: 100 per Task/Swap column (800 overall). RL gain: percentage points over SFT. Randomized SFT is rescored to the task-specific call budgets.}
    \label{tab:main-results}
    \footnotesize
    \setlength{\tabcolsep}{3pt}
    \begin{tabular*}{\textwidth}{@{\extracolsep{\fill}}ll rrrrrrrrrrr@{}}
        \toprule
        \multirow{3}{*}{Base model} & \multirow{3}{*}{Training} & \multicolumn{2}{c}{RoboTwin 2.0} & \multicolumn{9}{c}{LIBERO-Pro} \\
        \cmidrule(lr){3-4}\cmidrule(l){5-13}
        & & \multirow{2}{*}{Clean} & \multirow{2}{*}{Randomized} & \multicolumn{2}{c}{Spatial} & \multicolumn{2}{c}{Object} & \multicolumn{2}{c}{Goal} & \multicolumn{2}{c}{Long} & \multirow{2}{*}{Avg.} \\
        \cmidrule(lr){5-6}\cmidrule(lr){7-8}\cmidrule(lr){9-10}\cmidrule(lr){11-12}
        & & & & Task & Swap & Task & Swap & Task & Swap & Task & Swap & \\
        \midrule
        \multirow{3}{*}{Qwen3.5-9B} & Base & 20.0 & 15.6 & 1.0 & 1.0 & 2.0 & 0.0 & 5.0 & 5.0 & 0.0 & 0.0 & 1.8 \\
        & SFT & 31.2 & 26.4 & 77.0 & 71.0 & 86.0 & 79.0 & 52.0 & 49.0 & 19.0 & 29.0 & 57.8 \\
        & \textbf{SFT + RL} & \textbf{44.0} & \textbf{38.0} & \textbf{95.0} & \textbf{81.0} & \textbf{86.0} & \textbf{86.0} & \textbf{83.0} & \textbf{71.0} & \textbf{42.0} & \textbf{42.0} & \textbf{73.3} \\
        \midrule
        \multicolumn{2}{l}{\textbf{RL gain}} & \textbf{+12.8} & \textbf{+11.6} & \textbf{+18.0} & \textbf{+10.0} & \textbf{+0.0} & \textbf{+7.0} & \textbf{+31.0} & \textbf{+22.0} & \textbf{+23.0} & \textbf{+13.0} & \textbf{+15.5} \\
        \bottomrule
    \end{tabular*}
\end{table*}

\paragraph{Learning from teacher demonstrations.}
The teacher receives detailed tool descriptions, while the student uses LiteHarness's concise ones with the same tool names and argument schemas.
The teacher acts under the same memory rules, and each of its responses is paired with the context the student would receive at that call.
The policy is fine-tuned on successful demonstrations by maximizing the likelihood of teacher responses:
\begin{equation}
    \mathcal{L}_{\mathrm{SFT}}(\theta)
    = -\mathbb{E}_{(x,h,y^*)\sim\mathcal{D}_{\mathrm{SFT}}}
    \left[\log \pi_\theta(y^*\mid x,h)\right],
    \label{eq:sft}
\end{equation}
where $\mathcal{D}_{\mathrm{SFT}}$ contains task instructions $x$, student contexts $h$, and teacher responses $y^*$.

\paragraph{Multi-turn agentic reinforcement learning.}
The supervised policy is further trained with multi-turn group relative policy optimization (GRPO)~\cite{shao2024deepseekmath}.
In each round, a batch of tasks and initial states is selected by the schedule in Sec.~\ref{sec:task-aware-sampling}.
For each, the rollout policy $\pi_{\theta_{\mathrm{old}}}$ generates a group of $G$ trajectories through LiteHarness, which executes the requested tools and returns feedback until the episode ends.

Each trajectory $\tau_i$ receives a reward $r_i$ based on task success, with a bounded penalty for prolonged querying without robot action while the task remains incomplete.
Its group-relative advantage~\cite{feng2026group} is
\begin{equation}
    A_i = r_i - \frac{1}{G}\sum_{k=1}^{G}r_k.
    \label{eq:group-relative-advantage}
\end{equation}
Each response inherits its trajectory's group-relative advantage $A_i$, with a penalty $p_{i,t}$ subtracted for invalid tool calls.
To reduce the influence of trajectories with more training responses, we apply a weight $\bar M/M_i$, yielding $\widehat A_{i,t}=(\bar M/M_i)(A_i-p_{i,t})$.
Here, $M_i$ counts training responses in trajectory $\tau_i$, and $\bar M$ is the mean response count across trajectories in the batch.
With the token-level probability ratio
\begin{equation}
    \rho_{i,t,j}(\theta)
    = \frac{\pi_\theta(y_{i,t,j}\mid x_i,h_{i,t},y_{i,t,<j})}
    {\pi_{\theta_{\mathrm{old}}}(y_{i,t,j}\mid x_i,h_{i,t},y_{i,t,<j})},
    \label{eq:grpo-probability-ratio}
\end{equation}
where $y_{i,t,j}$ is the $j$-th token of response $y_{i,t}$, the policy minimizes
\begin{equation}
    \begin{aligned}
        \mathcal L_{\mathrm{GRPO}}(\theta)
        ={}& \mathbb E_{\mathcal B}\!\left[
        w_{i,t,j}\,\ell_{\mathrm{DC}}\!\left(
        \rho_{i,t,j}(\theta),\widehat A_{i,t}\right)\right] \\
        &{}-\beta\mathcal H(\pi_\theta),
    \end{aligned}
    \label{eq:multi-turn-grpo}
\end{equation}
where $\mathcal B$ indexes training tokens, $\ell_{\mathrm{DC}}$ is the dual-clipped loss, $w_{i,t,j}$ is the rollout correction weight, and $\beta$ weights the mean token entropy $\mathcal H$.
See Appendix~\ref{sec:app-training} for details.

\subsection{Interaction Budgets and Task Scheduling}
\label{sec:task-aware-sampling}

\paragraph{Interaction budgets.}
Each episode is limited by a model-call budget $B_{\mathrm{call}}$, which counts every model call including read-only queries, and a robot-action budget $B_{\mathrm{act}}$.
For each task, $B_{\mathrm{call}}$ is chosen from 80, 120, 160, and 200 calls so that it covers the longest successful teacher and SFT episodes.
$B_{\mathrm{act}}$ is 80, and an episode ends when either budget is exhausted or a terminal condition is reached.

\paragraph{Adaptive task scheduling.}
To improve training efficiency, we reduce sampling of tasks that consistently succeed or fail and allocate more rollouts to the remaining tasks.
Based on recent results, RoboHarness-RL applies three rules.
(1)~A task whose latest group fully succeeds skips one sampling cycle and is revisited before another skip, reducing sampling while maintaining coverage.
(2)~Tasks whose rollouts fail in two sampled groups are removed from the pool at a stage boundary.
(3)~On LIBERO-Pro, where SFT already solves most episodes of the Spatial and Object suites, later rounds draw twice as many groups from the harder Goal and Long suites.
Budget values and scheduling parameters are provided in Appendix~\ref{sec:app-budgets}.

\section{Experiments}
\label{sec:experiments}

\subsection{Experimental Setup}
\label{sec:experimental-setup}

\paragraph{Benchmark and protocol.}
We train a separate policy on each of RoboTwin 2.0~\cite{chen2025robotwin} and LIBERO-Pro~\cite{zhou2025libero}, with disjoint training and evaluation initial states.
RoboTwin has 50 bimanual tasks; we train only in clean scenes and evaluate on five held-out clean states per task with unseen instructions and on ORA's randomized set~\cite{ora2026code}, which adds distractors and varies backgrounds, lighting, and tabletop height (250 episodes each).
LIBERO-Pro covers the Spatial, Object, Goal, and Long suites under task and swap perturbations, with ten tasks per variant and ten initial states each (800 episodes).
Success follows the official task checks, and SFT and RL are evaluated on the same instances and settings.

\paragraph{Implementation details.}
Both policies use Qwen3.5-9B~\cite{qwen2026qwen35} and LiteHarness.
SFT uses 393 and 781 successful GPT-6 Astra demonstrations for RoboTwin and LIBERO-Pro, respectively.
We train for two epochs and initialize RL from the resulting SFT checkpoints.
Both stages use rank-16 LoRA with $\alpha=32$ and a frozen vision encoder, while perception tools and motion planners remain fixed.
Thinking is disabled throughout training and evaluation to keep multi-turn interaction affordable.
Each RL rollout batch samples eight trajectories for each of twelve task--initial-state pairs at temperature 1.0.
We report checkpoints after 25 sampling rounds on both benchmarks.
LiteHarness retains the current observation and up to three preceding observations in recent memory, with at most 24 images per model call.
Model-call budgets are set to 80--200 depending on the task, with a separate limit of 80 robot decisions per episode.
Training and evaluation use the same limits.
Evaluation uses temperature 0.7, top-$p$ 0.8, and top-$k$ 20.
All training runs on eight NVIDIA H200 GPUs.
Full training configurations are provided in Appendix~\ref{sec:app-training}.

\begin{table}[t]
    \centering
    \caption{LIBERO-Pro comparison. Suite success rates (\%) average Task and Swap; Avg. covers all 800 episodes.}
    \label{tab:comparison-libero}
    \fontsize{7.5}{9}\selectfont
    \setlength{\tabcolsep}{1pt}
    \begin{tabular*}{\columnwidth}{@{\extracolsep{\fill}}llcrrrrr@{}}
        \toprule
        System & Base model & VLA & Spat. & Obj. & Goal & Long & Avg. \\
        \midrule
        \multicolumn{8}{@{}l}{\textit{End-to-end VLAs}} \\
        $\pi_{0.5}$~\cite{zhang2026harness} & $\pi_{0.5}$ & -- & 10.5 & 9.0 & 20.0 & 4.5 & 11.0 \\
        $\pi_{\mathrm{RLinf}}$~\cite{zhang2026harness} & $\pi_{0.5}$ & -- & 50.5 & 74.5 & 43.5 & 31.5 & 50.0 \\
        \midrule
        \multicolumn{8}{@{}l}{\textit{Agents with frontier models via API}} \\
        Harness VLA~\cite{zhang2026harness} & GPT-6 Astra & Yes & 99.0 & 99.5 & 93.5 & 78.5 & 92.6 \\
        Harness VLA~\cite{zhang2026harness} & Opus 4.7 & Yes & 87.0 & 89.0 & 87.0 & 66.5 & 82.4 \\
        ORA~\cite{ora2026code} & DS-V4.1-Flash & No & 94.5 & 95.5 & 77.5 & 58.0 & 81.4 \\
        \midrule
        \multicolumn{8}{@{}l}{\textit{Agents with compact open-weight models}} \\
        AgenticRoboRL~\cite{liu2026agenticroborl} & Qwen3.5-9B & Yes & 91.0 & 95.5 & 57.5 & 40.0 & 71.0 \\
        \textbf{RoboHarness-RL} & \textbf{Qwen3.5-9B} & \textbf{No} & \textbf{88.0} & \textbf{86.0} & \textbf{77.0} & \textbf{42.0} & \textbf{73.3} \\
        \bottomrule
    \end{tabular*}
    \par\smallskip
    \parbox{\columnwidth}{\scriptsize DS: DeepSeek; VLA: learned VLA executor; --: inapplicable; bold: ours.}
\end{table}

\begin{figure}[t]
    \centering
    \includegraphics[width=\linewidth]{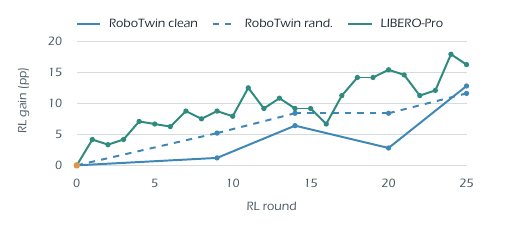}
    \caption{RL learning curves. Each point is the gain in success rate over SFT at an evaluated checkpoint (round 0: SFT). RoboTwin: full clean and randomized evaluations; LIBERO-Pro: initial states 1--3 of each task cell (240 episodes).}
    \label{fig:learning-curves}
\end{figure}

\begin{table}[t]
    \centering
    \caption{RoboTwin 2.0 success (\%). Data: post-pretraining robot data; ---: unavailable.}
    \label{tab:comparison}
    \fontsize{8.5}{10}\selectfont
    \setlength{\tabcolsep}{2.1pt}
    \begin{tabular*}{\columnwidth}{@{\extracolsep{\fill}}lll rr@{}}
        \toprule
        System & Base model & Data & Rand. & Clean \\
        \midrule
        \multicolumn{5}{@{}l}{\textit{End-to-end VLAs}$^{\dagger}$} \\
        $\pi_{0.5}$~\cite{intelligence2025pi_} & $\pi_{0.5}$ & RoboTwin & 46.0 & 70.7 \\
        X-VLA~\cite{robotwin2026leaderboard} & X-VLA & RoboTwin & 21.0 & 68.0 \\
        $\pi_0$~\cite{black2024pi_0} & $\pi_0$ & RoboTwin & 16.3 & 46.4 \\
        RDT~\cite{robotwin2026leaderboard} & RDT & RoboTwin & 13.7 & 34.5 \\
        \midrule
        \multicolumn{5}{@{}l}{\textit{Agents with frontier models via API}$^{\dagger}$} \\
        ORA~\cite{ora2026code} & DeepSeek-V4.1-Flash & -- & 69.2 & \textcolor{gray}{---} \\
        PyRUA-Lean$^{\ast}$~\cite{si2026fewer} & GPT-6 Astra & -- & 68.8 & \textcolor{gray}{---} \\
        RoboDawn~\cite{guo2026transferring} & GPT-6 Astra & -- & 53.2 & \textcolor{gray}{---} \\
        \midrule
        \multicolumn{5}{@{}l}{\textit{Agents with compact open-weight models}} \\
        K1~\cite{li2026robo} & Qwen3.5-9B & -- & 2.8 & 3.6 \\
        ORA~\cite{ora2026code} & Qwen3.5-9B & -- & 6.8 & 10.0 \\
        LiteHarness & Qwen3.5-27B & -- & 29.6 & 33.2 \\
        \textbf{RoboHarness-RL} & \textbf{Qwen3.5-9B} & \textbf{RoboTwin} & \textbf{38.0} & \textbf{44.0} \\
        \bottomrule
    \end{tabular*}
    \par\smallskip
    \parbox{\columnwidth}{\scriptsize $^{\dagger}$Leaderboard/paper results; $^{\ast}$no VLA skills.}
\end{table}

\subsection{Agentic RL beyond Imitation}
\label{sec:main-results}

\textbf{Agentic RL provides substantial gains beyond teacher imitation on both benchmarks} (Table~\ref{tab:main-results}).
We compare SFT and RL with the same Qwen3.5-9B backbone and LiteHarness, keeping evaluation conditions fixed within each benchmark.
Perception tools and motion planners remain unchanged, and RL uses no additional teacher demonstrations.
On RoboTwin 2.0, SFT raises clean-scene success from 20.0\% to 31.2\%, while subsequent RL brings it to 44.0\%.
The 12.8-point RL gain is comparable to the 11.2 points obtained from imitation, showing that teacher demonstrations leave considerable room for further improvement through interaction.
On LIBERO-Pro, success increases from 57.8\% after SFT to 73.3\% after RL, adding 15.5 percentage points across the full four-suite evaluation.
The gains build up over training: every evaluated checkpoint improves over SFT on both benchmarks (Fig.~\ref{fig:learning-curves}).

\textbf{The gains generalize to unseen scenes.}
Although RoboTwin training uses only clean scenes, RL raises randomized-scene success from 26.4\% to 38.0\%.
This 11.6-point gain is close to the 12.8 points in clean scenes, suggesting that the improvement extends beyond the visual conditions encountered during training.

\textbf{The gains are largest on the hardest suites.}
On LIBERO-Pro, Goal improves by 31.0 points under Task perturbations and 22.0 under Swap, and Long by 23.0 and 13.0 points.
Spatial and Object start from higher SFT baselines and gain 0.0--18.0 points, with Object under Task perturbations unchanged at 86.0\%.
Seven of the eight combinations of suite and perturbation type thus improve, while the magnitude depends on the task suite.

\begin{figure*}[t]
    \centering
    \includegraphics[width=\textwidth]{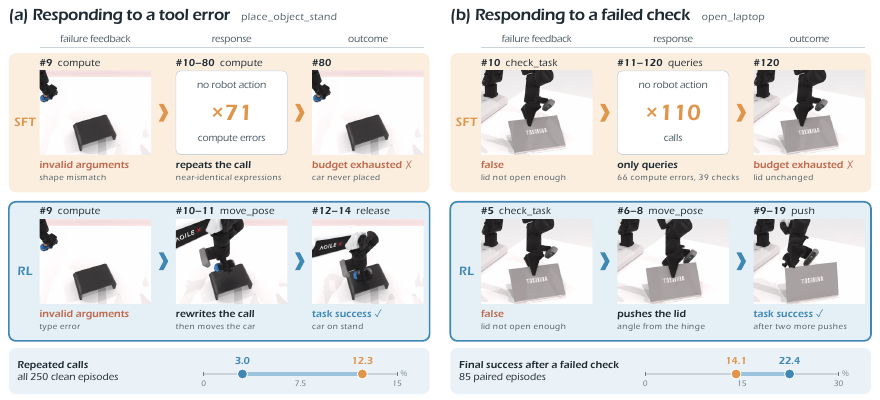}
    \caption{SFT/RL responses to invalid compute arguments (a) and negative completion checks (b). Frames replay actions; numbers index model calls. Bottom: repetition over 250 clean episodes and success over 85 pairs with negative checks for both policies.}
    \label{fig:feedback-execution}
\end{figure*}

\textbf{RL solves tasks that imitation does not.}
On LIBERO-Pro, the GPT-6 Astra teacher never completes ``put the left moka pot on the stove'' in 21 demonstration attempts, so SFT has no successful demonstration of this task and fails all ten evaluation episodes, whereas RL succeeds in seven.
More broadly, RL succeeds on nine tasks in each RoboTwin scene condition that SFT never solves in the evaluated trials.

\textbf{Improvements far outnumber regressions.}
On RoboTwin, success rises on 19 of 50 tasks and falls on nine in clean scenes, and rises on 21 and falls on seven in randomized scenes.
On LIBERO-Pro, comparing the two policies on the same 800 task--initial-state pairs, RL turns 155 SFT failures into successes, while 31 SFT successes become failures.
Regressions and remaining failures concentrate on long-horizon and precise manipulation: success stays limited on LIBERO-Pro Long, and RL fails all five clean three-block stacking trials, two of which SFT solves.

\subsection{Comparison with Existing Systems}
\label{sec:comparison}

\textbf{RoboHarness-RL is competitive with other compact open-weight systems} (Tables~\ref{tab:comparison-libero} and~\ref{tab:comparison}).
On LIBERO-Pro, our 9B model achieves 73.3\% overall success, outperforming the VLA-assisted AgenticRoboRL-9B by 2.3 percentage points in a more challenging setting: the VLM must specify and adjust robot actions through tool calls without a VLA or WAM.
The advantage is particularly pronounced on Goal, where our policy reaches 77.0\%, compared with 57.5\% for AgenticRoboRL.
Our policy also achieves 88.0\% on Spatial and 86.0\% on Object, supporting effective manipulation across both task and position perturbations with thinking mode disabled.
\begin{table}[t]
    \centering
    \caption{Mean counts per episode on states solved by both SFT and RL. Change: RL minus SFT; $^{*}$: $p<0.05$, paired sign test.}
    \label{tab:efficiency}
    \footnotesize
    \setlength{\tabcolsep}{1.5pt}
    \begin{tabular*}{\columnwidth}{@{\extracolsep{\fill}}l ccc ccc ccc@{}}
        \toprule
        & \multicolumn{3}{c}{VLM calls $\downarrow$} & \multicolumn{3}{c}{Read-only queries $\downarrow$} & \multicolumn{3}{c}{Robot decisions $\downarrow$} \\
        \cmidrule(lr){2-4} \cmidrule(lr){5-7} \cmidrule(l){8-10}
        & SFT & RL & Change & SFT & RL & Change & SFT & RL & Change \\
        \midrule
        RT clean & 23.3 & 15.4 & \textbf{$-$7.9}$^{*}$ & 12.7 & 7.3 & \textbf{$-$5.3}$^{*}$ & 8.1 & 5.8 & \textbf{$-$2.2}$^{*}$ \\
        RT rand. & 21.0 & 13.1 & \textbf{$-$7.9}$^{*}$ & 11.4 & 6.6 & \textbf{$-$4.7}$^{*}$ & 6.9 & 5.0 & \textbf{$-$1.8}\phantom{$^{*}$} \\
        LIBERO-Pro & 28.1 & 25.1 & \textbf{$-$2.9}$^{*}$ & 13.4 & 12.6 & \textbf{$-$0.8}\phantom{$^{*}$} & 13.1 & 11.4 & \textbf{$-$1.8}$^{*}$ \\
        \bottomrule
    \end{tabular*}
\end{table}

\textbf{On RoboTwin, policy training matters more than model scale.}
With the same LiteHarness, our trained 9B policy exceeds the untrained 27B model by 10.8 points in clean scenes and 8.4 in randomized scenes (Table~\ref{tab:comparison}).
The harness and training contribute separately: with the untrained 9B model, LiteHarness already reaches 20.0\% clean-scene success, compared with 10.0\% for ORA and 3.6\% for K1, and SFT and RL raise it to 44.0\%.

\textbf{A gap to frontier-model agents remains, mainly in randomized bimanual scenes.}
Against ORA with DeepSeek-V4.1-Flash, RoboHarness-RL is 8.1 points behind on LIBERO-Pro (73.3\% versus 81.4\%) but 31.2 points behind on randomized RoboTwin (38.0\% versus 69.2\%).

\subsection{Behavioral Analysis}
\label{sec:behavioral-analysis}

\textbf{RL reduces repetitive tool use.}
We count consecutive calls with identical arguments as repetitive when they cannot yield a new result: queries repeated without a state change, or actions retried after rejection or failure to move the robot.
Repeated actions that move the robot, such as successive pushes, are excluded.
Figure~\ref{fig:feedback-execution}a illustrates such repetition: SFT retries the same invalid compute call until its budget is exhausted.
RL reduces the repetition rate from 10.4\% to 2.8\% on clean RoboTwin, from 10.9\% to 2.5\% on randomized RoboTwin, and from 9.1\% to 5.7\% on LIBERO-Pro.
Even when all consecutive calls with identical arguments are counted, regardless of whether they change the environment, RL reduces repetition from 12.3\% to 3.0\% on clean RoboTwin, from 12.4\% to 2.7\% on randomized RoboTwin, and from 12.8\% to 7.5\% on LIBERO-Pro.

\textbf{RL completes tasks with fewer calls.}
We compare only initial states that both policies solve, since failed episodes mostly run until a budget is exhausted.
On these states, VLM calls per episode drop from 23.3 to 15.4 on clean RoboTwin, from 21.0 to 13.1 on randomized RoboTwin, and from 28.1 to 25.1 on LIBERO-Pro (Table~\ref{tab:efficiency}).
The savings come from different sources: on RoboTwin, mainly from fewer read-only queries (12.7 to 7.3 and 11.4 to 6.6), whereas on LIBERO-Pro query counts barely change and robot decisions drop from 13.1 to 11.4.
RL thus shortens successful executions by issuing fewer queries on RoboTwin and by needing fewer actions on LIBERO-Pro.

\textbf{RL recovers more often after negative feedback.}
Figure~\ref{fig:feedback-execution} compares their responses to negative feedback.
When a compute call is rejected for invalid arguments (a), SFT resubmits near-identical expressions until the budget runs out, whereas RL rewrites the call and completes the placement.
When a completion check reports that the laptop is not yet open (b), SFT keeps querying and checking, whereas RL pushes the lid further until the check succeeds.
When both policies receive execution-failure feedback on paired initial states, success rises from 23.4\% to 33.8\% on clean RoboTwin (201 pairs), 18.7\% to 30.6\% on randomized RoboTwin (209 pairs), and 42.5\% to 57.5\% on LIBERO-Pro (428 pairs), all significant (Appendix~\ref{sec:app-significance}).

\begin{figure}[t]
    \centering
    \includegraphics[width=\linewidth]{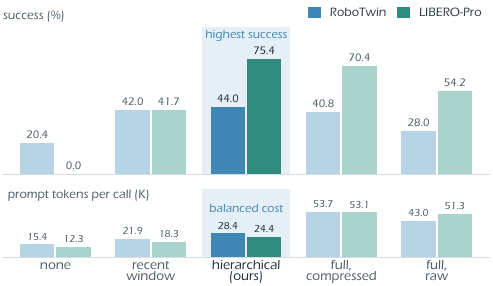}
    \caption{Memory ablation after RL. Hierarchical memory (shaded) gives the highest success at about half the tokens of compressed full history.}
    \label{fig:ablations}
\end{figure}

\subsection{Ablations}
\label{sec:efficiency-ablations}

\paragraph{Hierarchical memory.}
We vary only the visible interaction history, keeping prompts, tools, budgets, and initial states fixed (Fig.~\ref{fig:ablations}).
On LIBERO-Pro, we evaluate initial states 1--3 of each task cell (240 episodes).
For the RL model, hierarchical memory achieves 44.0\% success on RoboTwin and 75.4\% on LIBERO-Pro.
Removing history reduces success to 20.4\% and 0\%, respectively, while a recent window without summaries or a failure record reaches 42.0\% and 41.7\%.
Full raw history yields only 28.0\% success on RoboTwin and 54.2\% on LIBERO-Pro.
Even with ORA's built-in compression, full history uses roughly twice as many prompt tokens per call as hierarchical memory, yet achieves lower success rates: 40.8\% versus 44.0\% on RoboTwin and 70.4\% versus 75.4\% on LIBERO-Pro.

Since the RL policy is trained with hierarchical memory, its performance may partly reflect familiarity with this memory setting.
To examine whether the benefit extends to a model without such training, we also evaluate the base Qwen3.5-9B on RoboTwin.
Hierarchical memory achieves 20.0\% success, compared with 12.8\% without history and 8.4\% with the full raw history.
This consistent advantage suggests that the benefit is not solely due to training with hierarchical memory.

\textbf{These context savings also improve execution and training efficiency.}
Hierarchical memory uses about half the prompt tokens per call of the compressed full history (28.4K vs.\ 53.7K on RoboTwin and 24.4K vs.\ 53.1K on LIBERO-Pro) and 23\% less time per episode on RoboTwin.
On the same three RoboTwin task groups, one policy update under the compressed full history takes 1.65$\times$ the GPU time of hierarchical memory (4.00 vs.\ 2.43 GPU-hours).
Moreover, 76\% of its calls in the clean evaluation exceed our 45K-token training limit, and even with the limit raised to 78K, 10\% of its training rows still exceed it, whereas all hierarchical rows fit (Appendix~\ref{sec:app-compute}).

\paragraph{Task scheduling.}
\textbf{Adaptive scheduling sustains gains in later rounds} (Table~\ref{tab:sched-ablation}).\begin{table}[t]
    \centering
    \caption{Task-scheduling ablation on LIBERO-Pro (success, \%). Both runs continue from the round-15 checkpoint for ten rounds. T / S: task / swap variant (100 episodes each); All: 800 episodes.}
    \label{tab:sched-ablation}
    \footnotesize
    \setlength{\tabcolsep}{1.5pt}
    \begin{tabular*}{\columnwidth}{@{\extracolsep{\fill}}l cc cc cc cc c@{}}
        \toprule
        & \multicolumn{2}{c}{Spatial} & \multicolumn{2}{c}{Object} & \multicolumn{2}{c}{Goal} & \multicolumn{2}{c}{Long} & \\
        \cmidrule(lr){2-3} \cmidrule(lr){4-5} \cmidrule(lr){6-7} \cmidrule(lr){8-9}
        & T & S & T & S & T & S & T & S & All \\
        \midrule
        Round 15 & 88 & \textbf{82} & 86 & \textbf{86} & 71 & 67 & 39 & 33 & 69.0 \\
        Uniform & 87 & 78 & \textbf{99} & 82 & 74 & 65 & 33 & 38 & 69.5 \\
        Scheduled (ours) & \textbf{95} & 81 & 86 & \textbf{86} & \textbf{83} & \textbf{71} & \textbf{42} & \textbf{42} & \textbf{73.3} \\
        \bottomrule
    \end{tabular*}
\end{table}
Starting from the round-15 checkpoint (69.0\% success on 800 evaluation episodes), we compare ten additional rounds with adaptive scheduling versus uniform sampling.
Uniform sampling visits every task cell once per cycle in random order; both settings use the same learning rate, temperature, budgets, and batch size.
Adaptive scheduling reaches 73.3\%, significantly improving over the starting checkpoint ($p=0.002$), while uniform sampling remains at 69.5\% ($p=0.81$; paired sign tests).

Most of the advantage comes from Goal and Long, where adaptive scheduling produces 28 more successes than uniform sampling, compared with only two additional successes on Spatial and Object.
This matches the schedule's emphasis on the harder suites: from round 12 onward, Goal and Long receive eight of the twelve rollout groups per round.
The advantage persists across checkpoints: rounds 18, 20, and 24 achieve 577, 569, and 588 successes out of 800, respectively, all exceeding the uniform run's final 556.

\section{Conclusion}
\label{sec:conclusion}

RoboHarness-RL trains VLMs for tool-based manipulation through agentic RL, with LiteHarness supporting efficient multi-turn interaction.
RL improves success beyond teacher imitation without VLAs or WAMs, reduces repetitive calls, shortens successful executions, and improves recovery from negative feedback.
Gains extend to randomized scenes unseen during training.
Randomized bimanual manipulation and long-horizon tasks remain challenging; transfer to real robots remains to be evaluated.

\section*{LLM Usage Statement}
Our system uses Qwen3.5-9B as the policy and GPT-6 Astra for teacher demonstrations.
The authors used ChatGPT and Claude to edit the paper and improve code, verified all content, numbers, and claims, and take full responsibility for the paper.

{\small
\bibliographystyle{ieeenat_fullname}
\bibliography{main}
}
\clearpage
\appendix
\setcounter{figure}{0}
\setcounter{table}{0}
\renewcommand{\thefigure}{A\arabic{figure}}
\renewcommand{\thetable}{A\arabic{table}}
\maketitlesupplementary

\section*{Overview}
\label{sec:app-overview}

This supplementary material provides the details needed to interpret and reproduce the results in the main paper, together with additional analyses and qualitative examples.
It is organized as follows.

\begin{itemize}
\item \textbf{Appendix~\ref{sec:app-eval}: Evaluation Protocol and Full Results.}
Benchmark splits and scene perturbations, configurations of the compared systems, checkpoint selection, the evaluation runs underlying each reported number, per-task results on RoboTwin 2.0 and LIBERO-Pro, and statistical tests for the SFT--RL comparisons.
\item \textbf{Appendix~\ref{sec:app-harness}: LiteHarness.}
The tool interface, the student and teacher prompts, the complete student tool descriptions, and the hierarchical memory rules with a complete example context.
\item \textbf{Appendix~\ref{sec:app-sft}: Teacher Demonstrations and Supervised Fine-Tuning.}
Demonstration collection and filtering, dataset statistics, and the performance of the teacher.
\item \textbf{Appendix~\ref{sec:app-training}: Agentic RL Training.}
The training procedure, reward and advantage design, hyperparameters for both benchmarks, interaction budgets and task scheduling, compute, and training dynamics.
\item \textbf{Appendix~\ref{sec:app-qualitative}: Qualitative Examples.}
Complete SFT and RL trajectories with model responses and tool feedback, including recovery from execution failures and representative failure cases.
\item \textbf{Appendix~\ref{sec:app-limitations}: Limitations.}
Dependence on simulation, on the teacher, and on fixed tools, and the statistical scope of the results.
\end{itemize}

\section{Evaluation Protocol and Full Results}
\label{sec:app-eval}

\subsection{Benchmarks and Data Splits}
\label{sec:app-benchmarks}

We train a separate policy for each benchmark.
Table~\ref{tab:benchmark-splits} summarizes how initial states and instructions are divided between training and evaluation.

\begin{table}[h]
    \centering
    \caption{Training and evaluation splits. No evaluation initial state is used for teacher demonstrations or RL rollouts.}
    \label{tab:benchmark-splits}
    \footnotesize
    \setlength{\tabcolsep}{3pt}
    \begin{tabular*}{\columnwidth}{@{\extracolsep{\fill}}p{0.2\columnwidth}p{0.36\columnwidth}p{0.36\columnwidth}@{}}
        \toprule
        & Training & Evaluation \\
        \midrule
        RoboTwin, clean & Qualified \texttt{demo\_clean} seeds; task-level instructions & 5 held-out seeds per task (250); unseen instructions \\
        \addlinespace
        RoboTwin, randomized & -- & ORA's 5 released \texttt{demo\_randomized} seeds per task (250); unseen instructions \\
        \addlinespace
        LIBERO-Pro, Task / Swap & Initial states 11--49 & Initial states 1--10 of 80 task cells (800) \\
        \bottomrule
    \end{tabular*}
\end{table}

\paragraph{RoboTwin 2.0.}
RoboTwin 2.0~\cite{chen2025robotwin} contains 50 bimanual manipulation tasks, listed in Table~\ref{tab:per-task-results}.
An initial state is determined by the task and a random seed, which sets the object instances and their poses.
Following ORA~\cite{ora2026code}, we use only seeds that pass a qualification check: the benchmark's scripted expert solves the task, the success condition does not already hold at reset, and the final successful state remains stable for 250 further physics steps.
Success is determined by the official task-specific check.

Clean scenes use RoboTwin's \texttt{demo\_clean} configuration.
For each task, the first five qualified clean seeds that were not used in any earlier data collection form the held-out clean evaluation set, giving 250 episodes.
Each evaluation episode receives an instruction drawn from RoboTwin's unseen instruction split for the objects in that episode.\footnote{For \texttt{put\_object\_cabinet}, only four of the five evaluation seeds have the object information required to generate such an instruction; the fifth uses ORA's task-level instruction.}
The remaining qualified clean seeds are divided into disjoint sets for teacher demonstrations and RL rollouts; the RL pool contains 1{,}798 seeds, between 32 and 53 per task.
Training episodes use ORA's task-level instruction for each task, except for \texttt{place\_a2b\_left} and \texttt{place\_a2b\_right}, which have no task-level instruction and instead use instructions generated from the objects in each episode.

Randomized scenes use RoboTwin's \texttt{demo\_randomized} configuration, which places distractor objects on the table and randomizes background textures, lighting, and tabletop height.
We evaluate on the list of randomized episodes released by ORA, consisting of five seeds per task (250 episodes) with ORA's published unseen instructions.
No randomized scene is used during training.

\paragraph{LIBERO-Pro.}
LIBERO-Pro~\cite{zhou2025libero} extends LIBERO~\cite{liu2023libero} with perturbations that test whether a policy relies on memorized trajectories, using a single Franka Panda arm.
We use four suites, Spatial, Object, Goal, and Long (LIBERO-10), each with ten tasks, under two of the benchmark's perturbation types.
The Task variant modifies the target object or action of each task, introducing new goals that remain executable in the scene.
The Swap variant perturbs the initial configuration, altering the absolute and relative positions of objects while keeping the scene physically plausible; it corresponds to the position perturbation of LIBERO-Pro.
The four suites and two variants give 80 task cells, each a task under one variant, with 50 initial states.
Following AgenticRoboRL~\cite{liu2026agenticroborl}, initial states 1--10 of every cell form the evaluation set of 800 episodes, and states 11--49 are used for training, shared by teacher demonstrations and RL rollouts; state 0 is not used.
Each cell has a single instruction, so generalization on LIBERO-Pro is measured across unseen initial states rather than unseen instructions.
Success is determined by the official LIBERO success predicate.

\subsection{Compared Systems}
\label{sec:app-baselines}

\paragraph{Published results on RoboTwin 2.0.}
The VLA rows of Table~\ref{tab:comparison} are taken from the official RoboTwin 2.0 leaderboard~\cite{robotwin2026leaderboard}, updated on October 30, 2026.
These policies are trained on 50 clean demonstrations per task and evaluated with 100 trials per task and scene condition, so their evaluation instances differ from ours.
$\pi_{0.5}$ and X-VLA are co-trained across tasks, whereas $\pi_0$ and RDT are fine-tuned on each task separately.
For agents using frontier models through APIs, we report the randomized-scene results published for ORA~\cite{ora2026code}, RoboDawn~\cite{guo2026transferring} (zero-shot, Table~1 of the paper), and PyRUA-Lean~\cite{si2026fewer} (configuration without VLA skills, Table~3); none of them reports clean-scene results.

\paragraph{Baselines evaluated under our protocol.}
We evaluate three untrained baselines on the same clean and randomized RoboTwin episodes as RoboHarness-RL (Appendix~\ref{sec:app-benchmarks}), using the official success checks and a context window of 98{,}304 tokens.
\begin{itemize}
\item \textit{ORA with Qwen3.5-9B} uses the released ORA code with thinking disabled and the same task-specific budgets as LiteHarness.
It succeeds in 25/250 clean and 17/250 randomized episodes.
In clean scenes, 188 of the 250 episodes end because the context window is exhausted, after a median of 15--16 model calls, illustrating the context growth that hierarchical memory is designed to control.
\item \textit{K1 with Qwen3.5-9B}~\cite{li2026robo} retains its native configuration, including a limit of 400 model calls and temperature 0.2.
It succeeds in 9/250 clean and 7/250 randomized episodes.
In 130 of the 250 clean episodes, the arm never moves, because K1's grasp tool rejects target regions that are not elongated.
\item \textit{LiteHarness with Qwen3.5-27B} uses the same harness and budgets as our 9B policies.
It succeeds in 83/250 clean and 74/250 randomized episodes.
\end{itemize}

\paragraph{Published results on LIBERO-Pro.}
All rows of Table~\ref{tab:comparison-libero} other than ours are taken from the cited papers, and each suite score averages the reported Task and Swap results.
The $\pi_{0.5}$, $\pi_{\mathrm{RLinf}}$, and Harness VLA results come from Zhang~\etal~\cite{zhang2026harness}, with the Harness VLA rows taken from Tables~3 and~21 of that paper.
ORA with DeepSeek-V4.1-Flash reports 651 successes out of 800 episodes~\cite{ora2026code}.
For AgenticRoboRL~\cite{liu2026agenticroborl}, we use its RPent 9B variant trained with SFT and RL, without external memory (Table~1 of the paper); it executes actions through a learned VLA and uses the same split of initial states as ours.

\subsection{Checkpoint Selection}
\label{sec:app-checkpoints}

On both benchmarks, we report the checkpoint after 25 RL rounds.
Intermediate checkpoints are used to monitor training (Fig.~\ref{fig:learning-curves}) and in the scheduling ablation (Sec.~\ref{sec:efficiency-ablations}).
On RoboTwin, the checkpoints after rounds 9, 14, 20, and 25 are evaluated on the full clean set, and those after rounds 9, 13, 14, 20, and 25 on the full randomized set.
On LIBERO-Pro, every checkpoint up to round 25 is evaluated on initial states 1--3 of each task cell (240 episodes), and the checkpoints after rounds 11, 15, 17, 18, 20, 24, and 25 also on all 800 episodes.

\subsection{Evaluation Runs}
\label{sec:per-task-results}

\paragraph{Reported evaluations.}
Each number in Table~\ref{tab:main-results} comes from a single complete evaluation of each policy: 250 episodes per scene condition on RoboTwin and 800 episodes on LIBERO-Pro.
All paired analyses use these same evaluations, including the per-task results (Table~\ref{tab:per-task-results}), the behavioral statistics in Sec.~\ref{sec:behavioral-analysis}, and the statistics in Fig.~\ref{fig:feedback-execution}.
SFT and RL are therefore always compared on identical task--initial-state pairs.
Each checkpoint in Fig.~\ref{fig:learning-curves} is evaluated on the full clean and randomized sets on RoboTwin and on initial states 1--3 of each task cell on LIBERO-Pro (240 episodes).
Table~\ref{tab:main-results} reports the full 800-episode evaluation on LIBERO-Pro.

\paragraph{Sizes of the paired analyses.}
Statistics over all episodes use 250 pairs per RoboTwin condition and 800 pairs on LIBERO-Pro.
Both policies succeed on 51 clean and 42 randomized RoboTwin initial states and on 431 LIBERO-Pro initial states; these pairs underlie Table~\ref{tab:efficiency}.
Both policies receive execution-failure feedback in 201 clean, 209 randomized, and 428 LIBERO-Pro pairs, where success increases from 47 to 68, from 39 to 64, and from 182 to 246, respectively.
The statistic in Fig.~\ref{fig:feedback-execution}(b) uses the 85 clean pairs in which both policies receive at least one negative completion check, where success increases from 12 (14.1\%) to 19 (22.4\%).
On LIBERO-Pro, RL succeeds on 155 pairs where SFT fails, and SFT on 31 pairs where RL fails; the base model succeeds in 14 of the 800 episodes (1.75\%, shown as 1.8\% in Table~\ref{tab:main-results}).

\subsection{Per-Task Results}
\label{sec:app-per-task}

\begin{table*}[t]
\centering
\caption{Per-task success rates (\%) on RoboTwin under LiteHarness. Each entry uses the five episodes of the task in the evaluations of Table~\ref{tab:main-results}.}
\label{tab:per-task-results}
\scriptsize
\setlength{\tabcolsep}{3.2pt}
\begin{tabular*}{\textwidth}{@{\extracolsep{\fill}}l rrr rrr @{\hspace{10pt}} l rrr rrr@{}}
\toprule
 & \multicolumn{3}{c}{Clean} & \multicolumn{3}{c}{Randomized} & & \multicolumn{3}{c}{Clean} & \multicolumn{3}{c}{Randomized} \\
\cmidrule(lr){2-4}\cmidrule(lr){5-7}\cmidrule(lr){9-11}\cmidrule(lr){12-14}
Task & Base & SFT & RL & Base & SFT & RL & Task & Base & SFT & RL & Base & SFT & RL \\
\midrule
adjust\_bottle & 60 & 20 & 100 & 0 & 80 & 80 & place\_can\_basket & 0 & 40 & 40 & 0 & 20 & 20 \\
beat\_block\_hammer & 0 & 20 & 20 & 0 & 0 & 20 & place\_cans\_plasticbox & 0 & 40 & 0 & 0 & 20 & 40 \\
blocks\_ranking\_rgb & 0 & 60 & 40 & 0 & 40 & 40 & place\_container\_plate & 40 & 60 & 80 & 40 & 40 & 20 \\
blocks\_ranking\_size & 0 & 0 & 80 & 0 & 20 & 0 & place\_dual\_shoes & 0 & 0 & 0 & 0 & 0 & 0 \\
click\_alarmclock & 40 & 100 & 100 & 20 & 60 & 80 & place\_empty\_cup & 100 & 80 & 80 & 40 & 40 & 40 \\
click\_bell & 60 & 100 & 100 & 80 & 100 & 100 & place\_fan & 0 & 0 & 0 & 0 & 0 & 0 \\
dump\_bin\_bigbin & 0 & 0 & 20 & 0 & 0 & 0 & place\_mouse\_pad & 40 & 0 & 60 & 0 & 0 & 20 \\
grab\_roller & 20 & 80 & 100 & 0 & 60 & 100 & place\_object\_basket & 0 & 20 & 60 & 0 & 20 & 40 \\
handover\_block & 0 & 0 & 0 & 20 & 0 & 0 & place\_object\_scale & 20 & 40 & 40 & 40 & 60 & 80 \\
handover\_mic & 0 & 0 & 40 & 0 & 20 & 20 & place\_object\_stand & 80 & 20 & 60 & 0 & 20 & 60 \\
hanging\_mug & 0 & 20 & 20 & 0 & 0 & 0 & place\_phone\_stand & 0 & 0 & 0 & 0 & 0 & 60 \\
lift\_pot & 0 & 0 & 80 & 0 & 0 & 40 & place\_shoe & 0 & 0 & 0 & 40 & 20 & 20 \\
move\_can\_pot & 20 & 40 & 20 & 20 & 20 & 20 & press\_stapler & 100 & 100 & 100 & 100 & 100 & 100 \\
move\_pillbottle\_pad & 0 & 20 & 0 & 0 & 20 & 0 & put\_bottles\_dustbin & 0 & 20 & 0 & 0 & 20 & 20 \\
move\_playingcard\_away & 20 & 100 & 100 & 20 & 100 & 80 & put\_object\_cabinet & 0 & 0 & 0 & 0 & 0 & 0 \\
move\_stapler\_pad & 20 & 20 & 0 & 0 & 0 & 40 & rotate\_qrcode & 0 & 20 & 20 & 20 & 0 & 40 \\
open\_laptop & 40 & 20 & 40 & 80 & 20 & 40 & scan\_object & 20 & 0 & 0 & 0 & 0 & 0 \\
open\_microwave & 0 & 20 & 60 & 0 & 0 & 40 & shake\_bottle & 100 & 80 & 80 & 40 & 60 & 100 \\
pick\_diverse\_bottles & 0 & 0 & 20 & 0 & 0 & 40 & shake\_bottle\_horizontally & 60 & 80 & 60 & 80 & 80 & 80 \\
pick\_dual\_bottles & 0 & 0 & 0 & 0 & 20 & 0 & stack\_blocks\_three & 0 & 40 & 0 & 0 & 0 & 0 \\
place\_a2b\_left & 40 & 60 & 80 & 20 & 40 & 20 & stack\_blocks\_two & 0 & 0 & 80 & 20 & 40 & 80 \\
place\_a2b\_right & 20 & 60 & 100 & 0 & 60 & 60 & stack\_bowls\_three & 0 & 0 & 0 & 0 & 0 & 0 \\
place\_bread\_basket & 0 & 60 & 60 & 20 & 20 & 80 & stack\_bowls\_two & 0 & 0 & 0 & 0 & 20 & 20 \\
place\_bread\_skillet & 0 & 20 & 60 & 20 & 20 & 0 & stamp\_seal & 40 & 0 & 80 & 40 & 0 & 40 \\
place\_burger\_fries & 20 & 100 & 60 & 0 & 40 & 60 & turn\_switch & 40 & 0 & 60 & 20 & 20 & 60 \\
\midrule
\multicolumn{7}{l}{} & Mean (50 tasks) & 20.0 & 31.2 & 44.0 & 15.6 & 26.4 & 38.0 \\
\bottomrule
\end{tabular*}
\end{table*}

Table~\ref{tab:per-task-results} reports all 50 RoboTwin tasks for the three 9B training stages, using the evaluations in Table~\ref{tab:main-results}.
On clean scenes, RL improves over SFT on 19 tasks and falls behind on nine; nine tasks without any SFT success are solved at least once after RL.

\begin{table*}[t]
\centering
\caption{Per-task success rates (\%) on LIBERO-Pro. Each entry covers the ten evaluation initial states of a task; tasks are indexed by suite and task ID.}
\label{tab:per-task-libero}
\scriptsize
\setlength{\tabcolsep}{3.2pt}
\begin{tabular*}{\textwidth}{@{\extracolsep{\fill}}l rrr rrr @{\hspace{10pt}} l rrr rrr@{}}
\toprule
 & \multicolumn{3}{c}{Task} & \multicolumn{3}{c}{Swap} & & \multicolumn{3}{c}{Task} & \multicolumn{3}{c}{Swap} \\
\cmidrule(lr){2-4}\cmidrule(lr){5-7}\cmidrule(lr){9-11}\cmidrule(lr){12-14}
Task & Base & SFT & RL & Base & SFT & RL & Task & Base & SFT & RL & Base & SFT & RL \\
\midrule
Spatial 0 & 0 & 90 & 90 & 0 & 90 & 100 & Goal 0 & 0 & 0 & 0 & 0 & 50 & 50 \\
Spatial 1 & 0 & 90 & 100 & 0 & 90 & 80 & Goal 1 & 0 & 50 & 100 & 0 & 70 & 90 \\
Spatial 2 & 0 & 40 & 80 & 10 & 80 & 100 & Goal 2 & 10 & 30 & 70 & 0 & 70 & 100 \\
Spatial 3 & 0 & 80 & 100 & 0 & 90 & 100 & Goal 3 & 0 & 50 & 100 & 0 & 50 & 80 \\
Spatial 4 & 0 & 70 & 80 & 0 & 0 & 0 & Goal 4 & 0 & 30 & 100 & 0 & 80 & 90 \\
Spatial 5 & 0 & 80 & 100 & 0 & 80 & 100 & Goal 5 & 0 & 70 & 80 & 0 & 0 & 0 \\
Spatial 6 & 10 & 70 & 100 & 0 & 10 & 30 & Goal 6 & 10 & 60 & 80 & 0 & 0 & 80 \\
Spatial 7 & 0 & 90 & 100 & 0 & 100 & 100 & Goal 7 & 30 & 100 & 100 & 50 & 100 & 100 \\
Spatial 8 & 0 & 70 & 100 & 0 & 90 & 100 & Goal 8 & 0 & 70 & 100 & 0 & 10 & 20 \\
Spatial 9 & 0 & 90 & 100 & 0 & 80 & 100 & Goal 9 & 0 & 60 & 100 & 0 & 60 & 100 \\
\midrule
Object 0 & 0 & 100 & 100 & 0 & 90 & 100 & Long 0 & 0 & 90 & 100 & 0 & 80 & 70 \\
Object 1 & 0 & 90 & 100 & 0 & 100 & 100 & Long 1 & 0 & 20 & 20 & 0 & 60 & 70 \\
Object 2 & 0 & 100 & 100 & 0 & 10 & 10 & Long 2 & 0 & 20 & 80 & 0 & 60 & 100 \\
Object 3 & 10 & 70 & 100 & 0 & 40 & 50 & Long 3 & 0 & 0 & 10 & 0 & 0 & 0 \\
Object 4 & 0 & 100 & 100 & 0 & 100 & 100 & Long 4 & 0 & 0 & 0 & 0 & 0 & 20 \\
Object 5 & 0 & 100 & 100 & 0 & 90 & 100 & Long 5 & 0 & 10 & 60 & 0 & 10 & 70 \\
Object 6 & 0 & 100 & 100 & 0 & 80 & 100 & Long 6 & 0 & 20 & 0 & 0 & 0 & 0 \\
Object 7 & 0 & 90 & 60 & 0 & 90 & 100 & Long 7 & 0 & 30 & 80 & 0 & 50 & 40 \\
Object 8 & 10 & 20 & 10 & 0 & 100 & 100 & Long 8 & 0 & 0 & 70 & 0 & 30 & 50 \\
Object 9 & 0 & 90 & 90 & 0 & 90 & 100 & Long 9 & 0 & 0 & 0 & 0 & 0 & 0 \\
\midrule
\multicolumn{7}{l}{} & Mean (40 tasks) & 2.0 & 58.5 & 76.5 & 1.5 & 57.0 & 70.0 \\
\bottomrule
\end{tabular*}
\end{table*}

Table~\ref{tab:per-task-libero} reports the 40 LIBERO-Pro tasks under both variants, each evaluated on its ten evaluation initial states.
Across the 80 task cells, RL improves over SFT on 50, matches it on 24, and falls behind on 6, by at most three episodes.
RL succeeds on all ten initial states in 40 task cells, compared with 11 for SFT.
Four task cells without any SFT success are solved at least once after RL, including Goal~6 under Swap, which rises from 0 to 8 successes.
Eight task cells remain unsolved by both policies, five of them in the Long suite.

\subsection{Statistical Significance}
\label{sec:app-significance}

Because SFT and RL are evaluated on identical task--initial-state pairs, we compare them with paired tests (Table~\ref{tab:significance}).
For success rates, we report 95\% Wilson intervals for each policy and a 95\% interval for the RL gain from a bootstrap that resamples tasks (10{,}000 resamples), which accounts for the correlation among episodes of the same task.
The $p$-values come from exact two-sided McNemar tests on the discordant pairs.
The RL gain is significant on all three evaluations, and every bootstrap interval for the gain excludes zero.

\begin{table*}[t]
    \centering
    \caption{Paired comparison of SFT and RL success. Brackets: 95\% intervals (Wilson for success rates; task-level bootstrap for the gain). Discordant: pairs solved only by RL / only by SFT; $p$: exact McNemar test.}
    \label{tab:significance}
    \footnotesize
    \setlength{\tabcolsep}{4pt}
    \begin{tabular*}{0.8\textwidth}{@{\extracolsep{\fill}}lccccc@{}}
        \toprule
        & SFT (\%) & RL (\%) & Gain (pp) & Discordant & $p$ \\
        \midrule
        RT clean & 31.2 {\scriptsize[25.8, 37.2]} & 44.0 {\scriptsize[38.0, 50.2]} & 12.8 {\scriptsize[4.4, 21.6]} & 59 / 27 & $7{\times}10^{-4}$ \\
        RT rand. & 26.4 {\scriptsize[21.3, 32.2]} & 38.0 {\scriptsize[32.2, 44.2]} & 11.6 {\scriptsize[5.6, 18.0]} & 53 / 24 & $1{\times}10^{-3}$ \\
        LIBERO-Pro & 57.8 {\scriptsize[54.3, 61.1]} & 73.3 {\scriptsize[70.1, 76.2]} & 15.5 {\scriptsize[11.0, 20.2]} & 155 / 31 & $5{\times}10^{-21}$ \\
        \bottomrule
    \end{tabular*}
\end{table*}

The behavioral findings in Sec.~\ref{sec:behavioral-analysis} are tested with exact two-sided sign tests over paired episodes.
Per-episode rates of repetitive calls, as defined in Sec.~\ref{sec:behavioral-analysis}, are lower after RL in all three settings ($p=3{\times}10^{-7}$ and $4{\times}10^{-6}$ on clean and randomized RoboTwin, and $p=0.013$ on LIBERO-Pro).
Counting all identical consecutive calls gives $p<10^{-7}$ on both RoboTwin conditions and $p<10^{-16}$ on LIBERO-Pro.
On jointly successful initial states, RL uses fewer model calls in all three settings ($p=7{\times}10^{-5}$, $0.004$, and $0.010$ for clean RoboTwin, randomized RoboTwin, and LIBERO-Pro).
Read-only queries decrease significantly on both RoboTwin conditions ($p<0.001$) but not on LIBERO-Pro ($p=0.76$), consistent with the query counts reported there.
Robot decisions decrease significantly on clean RoboTwin ($p=0.02$) and LIBERO-Pro ($p=1{\times}10^{-4}$), but not on randomized RoboTwin ($p=0.15$).
Among pairs in which both policies receive execution-failure feedback, the increase in success is significant on clean RoboTwin ($p=0.013$), randomized RoboTwin ($p=0.002$), and LIBERO-Pro ($p=2{\times}10^{-10}$).
The narrower statistic in Fig.~\ref{fig:feedback-execution}(b), restricted to the 85 pairs with negative completion checks, points in the same direction but is not significant on its own ($p=0.23$).

The memory ablation (Fig.~\ref{fig:ablations}) is tested in the same way, pairing each variant with hierarchical memory on identical episodes.
On RoboTwin, the settings without history and with the full raw history are significantly lower, whereas the recent window and the compressed full history are not.

\section{LiteHarness}
\label{sec:app-harness}

\subsection{Tool Interface}
\label{sec:app-tools}

LiteHarness keeps the tools of ORA~\cite{ora2026code} and their argument schemas unchanged.
Table~\ref{tab:tools} lists the 20 tools available on RoboTwin.
SAM~3~\cite{carion2026sam} serves text grounding (\texttt{vision\_ground}) and point-prompted segmentation (\texttt{vision\_segment}).
Contact-GraspNet~\cite{sundermeyer2021contact} proposes grasps for \texttt{grasp\_plan}, combined on RoboTwin with a top-down grasp heuristic for the parallel-jaw grippers.
Motion targets are planned with cuRobo~\cite{sundaralingam2023curobo}, whereas \texttt{align\_plan} solves inverse kinematics only.
On LIBERO-Pro, the single-arm embodiment exposes 18 tools; \texttt{grasp\_execute} and \texttt{align\_plan} are absent, and motion tools take their targets without an arm argument.

Tools are either read-only or world-changing.
Read-only tools may be called several times or batched within one response.
A world-changing tool must be the only call in its response and ends the turn, after which the next observation shows its effect.
If a response mixes the two kinds, the read-only calls are executed and the world-changing call is rejected.
Each executed world-changing call counts toward the robot-action budget $B_{\mathrm{act}}$, and every model call counts toward $B_{\mathrm{call}}$.

\begin{table*}[t]
    \centering
    \caption{LiteHarness tools on RoboTwin. RO: read-only; WC: world-changing (the only call in its response; ends the turn). Optional arguments are omitted.}
    \label{tab:tools}
    \footnotesize
    \setlength{\tabcolsep}{4pt}
    \begin{tabular*}{\textwidth}{@{\extracolsep{\fill}}llp{0.25\textwidth}p{0.36\textwidth}c@{}}
        \toprule
        Tool & Type & Main arguments & Returns & Effect \\
        \midrule
        \texttt{observe} & Perception & -- & Images of the current or a saved observation & RO \\
        \texttt{vision\_ground} & Perception & camera, text & Ranked object candidates with 3D center, extent, and crop & RO \\
        \texttt{vision\_segment} & Perception & camera, pixel prompts & Object candidates with 3D center, extent, and crop & RO \\
        \texttt{mark\_point} & Perception & view, point names, pixels & 3D positions of the marked pixels and an overlay & RO \\
        \texttt{geometry\_query} & Perception & camera; world points, pixels, or region & Pixel or world coordinates, or depth statistics & RO \\
        \midrule
        \texttt{grasp\_plan} & Planning & arm, target reference & Grasp candidates with pre-grasp and contact poses, opening, and reachability & RO \\
        \texttt{align\_plan} & Planning & target reference, approach side, arm & Alignment pose and reachability & RO \\
        \texttt{compute} & Planning & expressions & Named numeric results & RO \\
        \midrule
        \texttt{grasp\_execute} & Action & candidate ID & Stage status; lift and opening in the next observation & WC \\
        \texttt{move\_to} & Action & target position per arm & Acknowledgement; motion receipt in the next observation & WC \\
        \texttt{move\_pose} & Action & target pose per arm & Compiled pose and reachability pre-check & WC \\
        \texttt{set\_gripper} & Action & open or close per arm & Submission; measured opening in the next observation & WC \\
        \texttt{rotate\_wrist} & Action & angle per arm & Acknowledgement & WC \\
        \texttt{rotate\_pitch} & Action & angle per arm & Acknowledgement & WC \\
        \texttt{recover\_pose} & Action & previous or home pose per arm & Acknowledgement & WC \\
        \texttt{wait\_ticks} & Action & number of ticks (1--20) & -- & WC \\
        \midrule
        \texttt{check\_task} & Episode & -- & Official task success (true or false) & RO \\
        \texttt{finish\_episode} & Episode & outcome, reason & Ends the episode & Terminal \\
        \texttt{report\_issue} & Episode & message & Recorded; the episode continues & RO \\
        \texttt{artifact\_read} & Memory & artifact reference & Bounded excerpt of a stored tool output & RO \\
        \bottomrule
    \end{tabular*}
\end{table*}

\subsection{Prompts and Tool Descriptions}
\label{sec:app-prompts}

The student's system prompt and rules are shown in Fig.~\ref{fig:student-prompt}.
The system prompt is shared by both benchmarks.
The rules differ only in the embodiment paragraph and in one recovery hint: the LIBERO-Pro version describes the single Franka Panda arm, its controlled point, the \texttt{agentview} and wrist cameras, and the command workspace, and it replaces the hint about moving the other arm with one about lifting after the jaw closes on an object.

Teacher and student share tool names and argument schemas but receive different descriptions.
The teacher sees ORA's full descriptions, while the student receives concise rewrites that keep the argument conventions, return fields, and common refusal reasons (Fig.~\ref{fig:tool-descriptions}).
On RoboTwin, the tool descriptions total 65.0K characters for the teacher and 16.7K for the student; on LIBERO-Pro, 55.3K and 15.1K.
Parameter descriptions in the student's schemas are shortened to at most 100 characters, while names, types, enumerations, and required fields are unchanged.
The first model call of an episode contains no interaction history.
In an example episode, this first call contains about 31K prompt tokens under ORA and 12.5K under LiteHarness.
The complete student descriptions of the RoboTwin tools are listed in Appendix~\ref{sec:app-tool-descriptions}.

For reference, the embodiment paragraph of the LIBERO-Pro rules reads:
\begin{quote}\footnotesize
Embodiment: fixed-base Franka Panda, one right 7-DoF arm with a parallel-jaw gripper. right\_grip\_site is the controlled point (already offset 0.097\,m from the end-effector root: do not add it again); every motion tool takes its fields at the top level, there is no arm choice. Positions are world metres, quaternions [x, y, z, w]. Local approach axis [0, 0, 1] (out of the gripper), jaw-closing line [1, 0, 0]; their current world directions are in the observation. Gripper opening is reported normalized (0 closed, 1 open), nominal maximum jaw separation 0.08\,m; a closed or stalled gripper does not by itself prove contact or retention. Finger collision extent along the approach axis from the controlled point is [$-$44, +9]\,mm. agentview is a fixed RGB-D camera; wrist rides on the arm and moves with it. Images are 256$\times$256, pixel origin top-left. Command workspace x [$-$0.6, 0.6], y [$-$0.5, 0.5], z [$-$0.1, 1.3].
\end{quote}
Its recovery hint for contact reads: ``blocked by contact $\rightarrow$ back off along the approach direction and come in from another side; a closed jaw that stalled on the object is the expected result of closing on it, then lift and observe.''

\begin{figure*}[t]
\centering
\fbox{\begin{minipage}{0.97\textwidth}\footnotesize
\textbf{System prompt (both benchmarks).}
You control one robot episode through the advertised tools. Use only the current observation, the tool results and the earlier-step summaries in this conversation. The Runtime executes actions and judges the task.
Turn protocol: read-only tools (grounding, planning, measuring, compute) may be called one after another or batched. A world-changing tool must be the only call in its reply; it ends the turn, and the next observation shows its effect. Every reply must contain a tool call.
Numbers you put into a tool argument must come from an observation, a tool result, or the compute tool; do not do arithmetic yourself. CANDIDATE images are previews, not world state. After an action, compare the before and after views before choosing the next tool; the latest observation is authoritative.
Earlier steps are summarized: one line per old step, then a list of the failed attempts so far, then the last three steps in full. Use finish\_episode only when its contract applies.

\medskip
\textbf{Rules (RoboTwin).}
\textit{Evidence:} a measurement or candidate is tied to the observation it came from. After any motion, geometry that could have moved is stale until a fresh observation or a new measurement confirms it; re-ground before planning on it. When two measurements disagree by enough to change the action, take another view or measurement; never average them. Command accepted, endpoint reached and task done are three different claims; only claim what the current evidence supports. An object missing from one camera may be out of its field or occluded: change the viewpoint instead of assuming it is gone. Depth-based world points come from the localization tools; calibration alone gives a ray.

\textit{When something fails:} a refusal, an error or an unchanged world says the approach was wrong, not only the tool. Before repeating, change what the attempt rests on: refresh the evidence the arguments came from, or take a different route to the same effect. Two failures of the same kind mean the premise is wrong. A failed attempt rules out that pose, contact or route, not the object or the goal. Typical recoveries: a stale candidate or reference $\rightarrow$ run grasp\_plan or vision\_ground again on the current observation; unreachable or planner failed $\rightarrow$ pick another candidate, use the reported reachable fraction or nearest reachable orientation, or lift first and move in a shorter segment; blocked by contact $\rightarrow$ back off along the approach direction and come in from another side or with the other arm; inter-arm collision $\rightarrow$ move the other arm away first.

\textit{Boundary:} the task sentence (and its public contract, when given) is the objective. Simulator state, segmentation truth, contacts and the success predicate are private. Only the evaluator's verdict at the end counts; motion completion is not success. Measured state can change without a command (settling, gravity).

\textit{Embodiment:} fixed-base ALOHA-AgileX, left and right 6-DoF arms with parallel-jaw grippers. left\_tcp and right\_tcp are the controlled points (already offset 0.12\,m from the flange: do not add it again). Positions are world metres, quaternions [x, y, z, w]. Local approach axis [1, 0, 0], jaw-closing line [0, 1, 0]; their current world directions are in the observation. Gripper opening is reported normalized (0 closed, 1 open), not as an aperture; a closed or stalled gripper does not by itself prove contact or retention. Finger collision extent along the approach axis from the TCP is [$-$50, +36]\,mm. head\_camera and front\_camera are fixed RGB-D; left\_camera and right\_camera ride on the arms. Images are 320$\times$240, pixel origin top-left.
\end{minipage}}
\caption{Student system prompt and rules on RoboTwin, reproduced verbatim.}
\label{fig:student-prompt}
\end{figure*}

\begin{figure*}[t]
\centering
\fbox{\begin{minipage}{0.97\textwidth}\footnotesize
\textbf{Teacher (ORA).}
Query the environment-owned official RoboTwin Boolean verdict at the current valid state boundary. Use when: fresh physical evidence suggests the requested arrangement may be complete, and always before finish\_episode(outcome=success); a planner success, closed gripper or visually plausible placement is not a substitute. Effect: read-only call to the trusted RoboTwin checker and no world advance. Returns: task\_success Boolean when available, or an explicit retryable unavailable result. Does not: explain the predicate, expose numerical tolerances, actor handles, target coordinates, task source or expert trajectory, change the world, or validate infrastructure quality. On error: Refused: any argument (invalid\_arguments). Wait for a valid state boundary and retry; never infer official success from action submission or visual appearance alone.

\medskip
\textbf{Student (LiteHarness).}
Ask the official RoboTwin checker whether the task is complete now. No arguments, read-only, no robot decision. Returns task\_success true or false (or a retryable unavailable result). Call it when fresh evidence suggests the arrangement is complete, and always before finish\_episode(success). A plausible-looking placement or a closed gripper is not success.
\end{minipage}}
\caption{Teacher and student descriptions of \texttt{check\_task}.}
\label{fig:tool-descriptions}
\end{figure*}

\subsection{Student Tool Descriptions}
\label{sec:app-tool-descriptions}

The following descriptions are given verbatim to the student policy on RoboTwin, in the order of Table~\ref{tab:tools}.
The LIBERO-Pro descriptions follow the same structure, adapted to the single arm and its workspace.

{\footnotesize\sloppy
\paragraph{\texttt{observe}.}
Return images of the current observation, or of a saved step of this round, without moving the world. The current RGB views and calibration are already delivered automatically after every action: call observe only for a named-point overlay (history\_\allowbreak{}point\_\allowbreak{}ids = solved mark names) or a saved step (step). views: subset of head\_\allowbreak{}camera, front\_\allowbreak{}camera, left\_\allowbreak{}camera, right\_\allowbreak{}camera (omit = every physical camera). Returns images, their dimensions and public\_\allowbreak{}robot\_\allowbreak{}facts (measured TCP poses, gripper command and measured opening). Refused: empty or repeated views, a step outside this round's saved steps, history\_\allowbreak{}point\_\allowbreak{}ids together with step. It never establishes grasp or task success.

\paragraph{\texttt{vision\_ground}.}
Find up to max\_\allowbreak{}candidates (1-3) regions matching a short visible description in text (at most 120 characters, e.g. 'green bottle', never a simulator name) in one current camera image. Returns ranked candidates with bbox\_\allowbreak{}xyxy, target\_\allowbreak{}ref, score, localization (center\_\allowbreak{}xyz\_\allowbreak{}m = median of the visible surface, support\_\allowbreak{}z\_\allowbreak{}m, mid\_\allowbreak{}height\_\allowbreak{}z\_\allowbreak{}m, height\_\allowbreak{}above\_\allowbreak{}support\_\allowbreak{}m) and numbered preview crops. Rank 0 and the score are not proof of identity: compare every crop with the task object (colour, shape, size, location) and reject one dominated by table, robot or background. For a parallel-jaw side grasp use mid\_\allowbreak{}height\_\allowbreak{}z\_\allowbreak{}m as the height reference, not center\_\allowbreak{}xyz\_\allowbreak{}m. Refused: empty text, text over 120 characters, max\_\allowbreak{}candidates outside 1-3, grounding service unavailable. target\_\allowbreak{}refs are bound to this observation: after any motion, ground again before grasp\_\allowbreak{}plan.

\paragraph{\texttt{vision\_segment}.}
Propose object masks around 1-8 independent pixel prompts on one camera image (current, or a saved step). Give 3-5 separated points inside the intended object; each point is answered on its own with up to max\_\allowbreak{}candidates (1-3) masks. Returns candidates[] (candidate\_\allowbreak{}id T0.., target\_\allowbreak{}ref, score, bbox\_\allowbreak{}xyxy, localization: center\_\allowbreak{}xyz\_\allowbreak{}m = median of the visible surface, support\_\allowbreak{}z\_\allowbreak{}m = height of the surface the object stands on, mid\_\allowbreak{}height\_\allowbreak{}z\_\allowbreak{}m, height\_\allowbreak{}above\_\allowbreak{}support\_\allowbreak{}m) and preview crops. Choose the candidate whose crop is the intended object; the score orders the masks of one point only. A point on background, an edge or a shadow can return the table or a neighbour. Refused: a pixel outside the image (whole call), more than 8 points, max\_\allowbreak{}candidates outside 1-3. target\_\allowbreak{}refs are bound to this observation: after any motion, segment again before grasp\_\allowbreak{}plan.

\paragraph{\texttt{mark\_point}.}
Measure 1-8 pixels of one current RGB view as world points in metres. view = the camera the pixels were read from; point\_\allowbreak{}ids and points\_\allowbreak{}uv are index-aligned and of equal length; each pixel is [u,\allowbreak{}v] in integer pixels of that view (320x240, origin top-left), well inside the visible surface, not on an edge or shadow. Returns one marks[] entry per pixel (status, xyz\_\allowbreak{}m, depth-support counts) and one overlay image. A name already solved from another view is verified against the stored point: a disagreement over 25 mm fails that entry (inconsistent\_\allowbreak{}mark) and keeps the stored one. Per-entry failures: depth\_\allowbreak{}discontinuity or insufficient\_\allowbreak{}depth\_\allowbreak{}support $\rightarrow$ aim farther inside the surface; invalid\_\allowbreak{}depth $\rightarrow$ another pixel. The whole call is refused for more than 8 entries, unequal lengths, repeated or empty names, or negative pixels. A wrist camera moves with its arm: after any motion measure again from the fresh observation. A measured pixel is a point on the visible surface, not the object centre.

\paragraph{\texttt{geometry\_query}.}
Numbers only, from one camera's published RGB-D and calibration: no image, no mark. Send camera plus exactly ONE of: world\_\allowbreak{}points\_\allowbreak{}m (1-16 [x,\allowbreak{}y,\allowbreak{}z] metres $\rightarrow$ pixel\_\allowbreak{}uv, in\_\allowbreak{}front, in\_\allowbreak{}frame, scene depth at that pixel and depth\_\allowbreak{}residual\_\allowbreak{}m; a positive residual means the visible surface is nearer than your point, i.e. it is probably occluded), points (1-16 [u,\allowbreak{}v] pixels $\rightarrow$ world\_\allowbreak{}xyz\_\allowbreak{}m from the median depth in a window of half-width window\_\allowbreak{}px, default 3, range 0-15), or region ([x0,\allowbreak{}y0,\allowbreak{}x1,\allowbreak{}y1] with x0\textless{}x1, y0\textless{}y1 $\rightarrow$ valid depth count and fraction, min/\allowbreak{}median/\allowbreak{}max depth). Optional step reads a saved observation. Refused: none or two of the three operations, a pixel or rectangle outside the image, more than 16 points, window\_\allowbreak{}px outside 0-15. Missing depth is reported as valid false, not an error: move the pixel inside the surface or widen window\_\allowbreak{}px. Use mark\_\allowbreak{}point when the pixel should be kept under a name and cross-checked; use compute for arithmetic on numbers you already hold.

\paragraph{\texttt{grasp\_plan}.}
Generate up to max\_\allowbreak{}candidates (1-3) parallel-jaw grasp candidates for a target\_\allowbreak{}ref from vision\_\allowbreak{}ground or vision\_\allowbreak{}segment on the CURRENT observation, for the chosen arm. Returns candidates with candidate\_\allowbreak{}id, world-frame pregrasp\_\allowbreak{}pose and contact\_\allowbreak{}pose (position\_\allowbreak{}m, orientation\_\allowbreak{}xyzw), predicted opening, ik\_\allowbreak{}reachability \{reachable, pregrasp, approach, approach\_\allowbreak{}tilt\_\allowbreak{}from\_\allowbreak{}vertical\_\allowbreak{}deg\} and review images; reachable candidates and near-vertical approaches are listed first, but rank is not a success probability: inspect the panels. Then either call grasp\_\allowbreak{}execute(candidate\_\allowbreak{}id, rationale) for the committed open/\allowbreak{}pregrasp/\allowbreak{}approach/\allowbreak{}close/\allowbreak{}lift sequence, or copy a pose into move\_\allowbreak{}pose under targets.\textless{}arm\textgreater{} with position\_\allowbreak{}mode=absolute, orientation\_\allowbreak{}mode=absolute\_\allowbreak{}quaternion and position\_\allowbreak{}m / orientation\_\allowbreak{}xyzw unchanged when you need a decision before closing. Any world action stales the candidates: run grasp\_\allowbreak{}plan again on the new observation before executing. Refused: target\_\allowbreak{}ref unknown (target\_\allowbreak{}ref\_\allowbreak{}unavailable) or from an earlier observation (stale\_\allowbreak{}observation: re-ground first); backend failure (grasp\_\allowbreak{}proposal\_\allowbreak{}failed: retry once, then another view). status no\_\allowbreak{}candidate: every candidate needed an opening over 94.9 mm or lay outside the workspace; take a better view or choose another target.

\paragraph{\texttt{align\_plan}.}
Inverse-kinematics-only check of standoff poses facing a current target\_\allowbreak{}ref from one world side: top (above the target, travelling -Z), front (at -X travelling +X), back (at +X), left (at +Y), right (at -Y); approach\_\allowbreak{}side names a world direction, never an arm. standoff\_\allowbreak{}m 0.02-0.30 (default 0.10) from the target's visible-surface centroid; max\_\allowbreak{}candidates (1-6) jaw rolls; arm left, right or both. Returns candidates with align\_\allowbreak{}pose \{position\_\allowbreak{}m, orientation\_\allowbreak{}xyzw\}, reachable, position\_\allowbreak{}error\_\allowbreak{}m, orientation\_\allowbreak{}error\_\allowbreak{}deg. No motion is submitted and nothing is collision checked: a reachable pose can still fail in move\_\allowbreak{}pose, and an unreachable one only describes this arm from its present joints. Optional diagnosis; grasp\_\allowbreak{}plan does not use it. Refused: stale or unknown target\_\allowbreak{}ref, values outside the ranges.

\paragraph{\texttt{compute}.}
Evaluate a short block of numeric Python on numbers you already hold. Call it before using any derived number (offset, distance, angle, quaternion, unit conversion); never do mental arithmetic. Arguments: bindings = list of \{name, value\} (a number, boolean, vector or matrix; lists become numpy arrays, so b - a works); expressions = ONE string of straight-line statements separated by \textbackslash{}n (no Markdown fences); outputs = the final variable names to return, each one assigned in the block (the last bare expression is also available as result).

Prebound names, never import or rebind them: np, Rotation (scipy; quaternions are [x,\allowbreak{}y,\allowbreak{}z,\allowbreak{}w]), math, pi, float, int, abs, max, min, round, and geom with four helpers: geom.unit(v) normalizes a 3-vector; geom.angle\_\allowbreak{}deg(a, b) is the unsigned angle 0-180; geom.rotate(q, v) rotates v by quaternion q; geom.transform\_\allowbreak{}point(t, q, p) returns t + rotated p.

Correct forms: np.linalg.norm(b - a); np.cross(a, b) (a function: a list or array has no .cross); np.clip(x, -1, 1); Rotation.from\_\allowbreak{}quat(q).as\_\allowbreak{}matrix() (q has 4 components); Rotation.from\_\allowbreak{}euler('xyz', [r, p, y], degrees=True).as\_\allowbreak{}quat(); Rotation.from\_\allowbreak{}rotvec(geom.unit(axis) * math.radians(deg)).as\_\allowbreak{}quat(); (Rotation.from\_\allowbreak{}quat(q2) * Rotation.from\_\allowbreak{}quat(q1)).as\_\allowbreak{}quat() applies q1 then q2; math.radians(deg), math.degrees(rad).

Examples: \{"bindings":\allowbreak{}[\{"name":\allowbreak{}"a",\allowbreak{}"value":\allowbreak{}[0.1,\allowbreak{}0.2,\allowbreak{}0.3]\},\allowbreak{}\{"name":\allowbreak{}"b",\allowbreak{}"value":\allowbreak{}[0.4,\allowbreak{}0.6,\allowbreak{}0.3]\}],\allowbreak{}"expressions":\allowbreak{}"distance\_\allowbreak{}m = np.linalg.norm(b - a)",\allowbreak{}"outputs":\allowbreak{}["distance\_\allowbreak{}m"]\} returns 0.5. \{"bindings":\allowbreak{}[\{"name":\allowbreak{}"p",\allowbreak{}"value":\allowbreak{}[0.1,\allowbreak{}0.2,\allowbreak{}0.3]\},\allowbreak{}\{"name":\allowbreak{}"q",\allowbreak{}"value":\allowbreak{}[0,\allowbreak{}0,\allowbreak{}0,\allowbreak{}1]\}],\allowbreak{}"expressions":\allowbreak{}"axis = geom.rotate(q, [1, 0, 0])\textbackslash{}npoint = p + 0.03 * axis\textbackslash{}nangle = geom.angle\_\allowbreak{}deg(axis, [0, 0, 1])",\allowbreak{}"outputs":\allowbreak{}["point",\allowbreak{}"angle"]\} returns point [0.13,\allowbreak{}0.2,\allowbreak{}0.3] and angle 90.

Limits: at most 24 statements, 16 bindings, 12 outputs, 64 elements per value; no loops, comprehensions, if statements, def, lambda, import, print, return, dict literals, augmented assignment (x += 1) or string operations. Each call starts fresh: pass every input again in bindings. A refusal is invalid\_\allowbreak{}arguments with field, received and constraints: a Python AttributeError or ValueError in the message means the code itself is wrong (fix the call form), an unknown name means a missing binding, an output name must be assigned in the block. Correct the block; never resend it unchanged.

\paragraph{\texttt{grasp\_execute}.}
Run one committed grasp as one robot decision: open the gripper (this releases anything it holds), move to the candidate's pregrasp pose, approach to the contact pose, close, lift 3 cm, verify. candidate\_\allowbreak{}id must come from grasp\_\allowbreak{}plan on the current observation; rationale is one sentence of at most 200 characters. Before moving, both poses get a read-only IK and collision probe; a grasp\_\allowbreak{}pregrasp\_\allowbreak{}unreachable refusal submits nothing: pick another candidate or move the arm to a better configuration first. The next observation reports the stage statuses, the measured lift and the jaw opening: command close with measured\_\allowbreak{}opening\_\allowbreak{}normalized between about 0.05 and 0.95 suggests something is between the jaws (evidence, not proof). A stage that does not arrive ends the sequence there. Refused: stale\_\allowbreak{}candidate\_\allowbreak{}observation (candidate from an earlier observation: run grasp\_\allowbreak{}plan again), candidate\_\allowbreak{}unavailable, another sequence in flight. Never call it while already holding the object you need: the opening stage drops it. Use move\_\allowbreak{}pose plus set\_\allowbreak{}gripper instead for a grasp that must not lift, or for handles and articulated parts.

\paragraph{\texttt{move\_to}.}
Translate one or both TCPs to absolute world positions while holding each arm's current orientation, routed through the collision-aware global planner. targets \{left\textbar{}right: \{position\_\allowbreak{}m: [x,\allowbreak{}y,\allowbreak{}z] metres\}\} and nothing else (no orientation fields: use move\_\allowbreak{}pose for those). Refused before submission: a target equal to the current position, a position outside the command workspace x [-0.6,\allowbreak{}0.6], y [-0.4,\allowbreak{}0.4], z [0.7,\allowbreak{}1.45] (workspace\_\allowbreak{}bounds). The acknowledgement carries reach\_\allowbreak{}along\_\allowbreak{}request (approximate IK pre-check) and observed\_\allowbreak{}clearance; the receipt in the next observation says reached, planner\_\allowbreak{}failed, plan\_\allowbreak{}rejected or not\_\allowbreak{}reached with residuals. Do not resend a failed target unchanged.

\paragraph{\texttt{move\_pose}.}
Command a full TCP pose for one or both arms under targets.\textless{}left\textbar{}right\textgreater{}. Each arm names position\_\allowbreak{}mode and orientation\_\allowbreak{}mode and sends exactly that mode's fields:

position\_\allowbreak{}mode absolute $\rightarrow$ position\_\allowbreak{}m [x,\allowbreak{}y,\allowbreak{}z] world metres; delta $\rightarrow$ translation\_\allowbreak{}m [dx,\allowbreak{}dy,\allowbreak{}dz] metres + position\_\allowbreak{}frame (world \textbar{} end\_\allowbreak{}effector); hold\_\allowbreak{}current $\rightarrow$ no position fields.

orientation\_\allowbreak{}mode absolute\_\allowbreak{}quaternion $\rightarrow$ orientation\_\allowbreak{}xyzw [x,\allowbreak{}y,\allowbreak{}z,\allowbreak{}w] unit quaternion (norm 0.98-1.02); world\_\allowbreak{}axes $\rightarrow$ approach\_\allowbreak{}world (world direction the jaws point along, TCP local +X) + jaw\_\allowbreak{}line\_\allowbreak{}world (jaw closing line, local +Y); axis\_\allowbreak{}angle\_\allowbreak{}delta $\rightarrow$ rotation\_\allowbreak{}axis (x\textbar{}y\textbar{}z) + rotation\_\allowbreak{}angle\_\allowbreak{}deg (signed DEGREES, -180..180) + orientation\_\allowbreak{}frame (world \textbar{} end\_\allowbreak{}effector); hold\_\allowbreak{}current $\rightarrow$ no orientation fields.

interaction: free\_\allowbreak{}space (default; global planner), precision (global planner plus a small local settle; use for placing), contact\_\allowbreak{}approach (local push of up to 10 mm / 5 deg; a longer approach plans to a pre-contact pose first). Two arms in one call must share one interaction.

Example: \{"targets":\allowbreak{}\{"right":\allowbreak{}\{"position\_\allowbreak{}mode":\allowbreak{}"absolute",\allowbreak{}"position\_\allowbreak{}m":\allowbreak{}[0.1,\allowbreak{}0.0,\allowbreak{}0.9],\allowbreak{}"orientation\_\allowbreak{}mode":\allowbreak{}"absolute\_\allowbreak{}quaternion",\allowbreak{}"orientation\_\allowbreak{}xyzw":\allowbreak{}[0,\allowbreak{}0,\allowbreak{}0,\allowbreak{}1],\allowbreak{}"interaction":\allowbreak{}"precision"\}\}\}. Example lift: \{"targets":\allowbreak{}\{"left":\allowbreak{}\{"position\_\allowbreak{}mode":\allowbreak{}"delta",\allowbreak{}"translation\_\allowbreak{}m":\allowbreak{}[0,\allowbreak{}0,\allowbreak{}0.05],\allowbreak{}"position\_\allowbreak{}frame":\allowbreak{}"world",\allowbreak{}"orientation\_\allowbreak{}mode":\allowbreak{}"hold\_\allowbreak{}current"\}\}\}.

Returns immediately status submitted with compiled\_\allowbreak{}pose: target\_\allowbreak{}pose\_\allowbreak{}xyz\_\allowbreak{}xyzw, target\_\allowbreak{}axes\_\allowbreak{}world, reach\_\allowbreak{}along\_\allowbreak{}request (approximate IK pre-check: target\_\allowbreak{}reachable, and when false the farthest reachable fraction of the straight path and nearest\_\allowbreak{}reachable\_\allowbreak{}orientation) and observed\_\allowbreak{}clearance (lowest point of the jaws or held object at the target minus the highest depth surface under it; negative means that surface is higher). The next observation's receipt says reached, or planner\_\allowbreak{}failed (IK\_\allowbreak{}FAIL: no collision-free plan; try a nearer intermediate pose, the nearest reachable orientation or the other arm), plan\_\allowbreak{}rejected (workspace or inter-arm clearance: move the other arm away first), not\_\allowbreak{}reached with blocked (stalled on contact: back off along the approach direction or recover\_\allowbreak{}pose). Refused before submission: an unknown field, a mode without its fields or a field of another mode, hold\_\allowbreak{}current for both halves, a target equal to the measured pose, a position outside the command workspace x [-0.6,\allowbreak{}0.6], y [-0.4,\allowbreak{}0.4], z [0.7,\allowbreak{}1.45] (workspace\_\allowbreak{}bounds). Numbers must come from measurements or compute. Read the fresh observation before the next action; do not resend the same failed target.

\paragraph{\texttt{set\_gripper}.}
Open or close one or both grippers in one call: targets \{left: open\textbar{}close, right: open\textbar{}close\}; an omitted gripper keeps its current directive. One robot decision; the arm poses are held. Optional target\_\allowbreak{}ref is only your note of what you are closing on. Returns status submitted; the next observation gives the measured opening (0 closed, 1 open). A closed jaw is not proof of a grasp; opening releases whatever is held. To grasp a planned candidate prefer grasp\_\allowbreak{}execute. Refused: empty targets, an arm other than left/\allowbreak{}right, a directive other than open/\allowbreak{}close.

\paragraph{\texttt{rotate\_wrist}.}
Roll one or both TCPs about their own approach axis (local +X) by signed degrees while holding position: targets \{left\textbar{}right: degrees in [-180,\allowbreak{}180], not 0\}. Changes only the jaw line. Up to 15 deg is local control; larger rolls use the global planner. Refused while that TCP is outside the command workspace. Use move\_\allowbreak{}pose when position must change too.

\paragraph{\texttt{rotate\_pitch}.}
Tilt one or both TCPs about their jaw axis (local +Y) by signed degrees while holding position: targets \{left\textbar{}right: degrees in [-180,\allowbreak{}180], not 0\}. Changes the approach direction; the resulting world direction depends on the current orientation (+90 is not always world-down), so read target\_\allowbreak{}axes\_\allowbreak{}world in the acknowledgement. Up to 15 deg is local; larger tilts use the global planner. Refused while outside the command workspace.

\paragraph{\texttt{recover\_pose}.}
Plan one or both arms to a safe joint configuration, keeping every gripper command: targets \{left\textbar{}right: previous \textbar{} home\}. previous = the last safely reached configuration of a successful pose action; home = the task's start joints. Use after a blocked or failed motion (after a blocked failure it first retreats 20 mm opposite the recorded approach); prefer one arm at a time. It does not restore moved objects or undo anything in the scene.

\paragraph{\texttt{wait\_ticks}.}
Hold every command and let simulated time pass: ticks 1-20, one tick = 40 ms. One robot decision. Use when a released object is settling or a scripted scene animation is running; not a fix for a pose residual (move\_\allowbreak{}pose) or a stalled jaw (set\_\allowbreak{}gripper).

\paragraph{\texttt{check\_task}.}
Ask the official RoboTwin checker whether the task is complete now. No arguments, read-only, no robot decision. Returns task\_\allowbreak{}success true or false (or a retryable unavailable result). Call it when fresh evidence suggests the arrangement is complete, and always before finish\_\allowbreak{}episode(success). A plausible-looking placement or a closed gripper is not success.

\paragraph{\texttt{finish\_episode}.}
End the episode. outcome: success (only after check\_\allowbreak{}task returned true at the current state), give\_\allowbreak{}up (requires category: unreachable \textbar{} object\_\allowbreak{}lost \textbar{} scene\_\allowbreak{}damaged \textbar{} repeated\_\allowbreak{}failure \textbar{} other) or abort (only for an infrastructure failure the harness recorded; planner failures, stalls, refused calls and check\_\allowbreak{}task false are NOT infrastructure). reason: at least 24 characters and 4 words; for success cite the evidence and its source (check\_\allowbreak{}task, an image, a measured value), for give\_\allowbreak{}up state what was observed and attempted. Refused: task\_\allowbreak{}success\_\allowbreak{}not\_\allowbreak{}confirmed, finish\_\allowbreak{}category\_\allowbreak{}required, finish\_\allowbreak{}reason\_\allowbreak{}insufficient, action\_\allowbreak{}pending\_\allowbreak{}fresh\_\allowbreak{}observation (wait for the observation first).

\paragraph{\texttt{report\_issue}.}
Record one concrete tool or interface defect (message): a malformed result, a stale binding, contradictory camera evidence. No robot decision, the episode continues, nothing is repaired. Ordinary task difficulty is not an issue.

\paragraph{\texttt{artifact\_read}.}
Read exact text or JSON behind an artifact:\allowbreak{}/\allowbreak{}/ reference from a \textless{}recoverable\_\allowbreak{}artifact\textgreater{} pointer: ref plus either a byte range (offset, limit up to 8192) or a json\_\allowbreak{}pointer. Images are reopened with observe(ref=...) instead.

}

\subsection{Hierarchical Memory}
\label{sec:app-memory}

LiteHarness builds each context from the recorded interaction by deterministic rules, without additional model calls.
The same rules render teacher recordings into student contexts for SFT.

\paragraph{Recent memory $R_t$.}
The window contains the current step and the $K=3$ preceding steps, where a step is the interaction between two consecutive observations.
Each observation's text body is shortened to at most 2{,}500 characters; image retention follows the separate rule below.
Robot actions and their results are kept, with numbers rounded to four decimals.
Within each step, the last two read-only queries are kept exactly.
Earlier queries are compressed: the model's text and the call arguments are shortened to at most 300 characters, and each result is replaced by a digest of at most 900 characters.
The digest keeps the fields needed for later tool use, including status and failure codes, candidate IDs and target references, scores, 3D centers and extents, bounding boxes, support heights, positions, orientations, and task success, with at most eight items per list.

\paragraph{Historical summaries $S_t$.}
Steps older than the window are reduced to their action.
The observation is replaced by a single line naming the last action and its outcome, the action call is kept with its arguments shortened to 300 characters, and its result is replaced by a digest.
Read-only queries of these steps are removed.

\paragraph{Failure record $F_t$.}
The failure record lists every failed action and every rejected call from completed steps, with the step index, the tool, compressed arguments, and the reported cause.
It keeps the 20 most recent entries and is placed immediately before the steps shown in full.
Figure~\ref{fig:memory-format} shows the formats of the step summaries and the failure record.

\paragraph{Images.}
Images are retained only from the current step and the two preceding steps, a shorter window than the textual recent memory.
Images from older steps are omitted even when their text remains in $R_t$.
Images returned by tools before the current observation are replaced by a placeholder, so tool-generated previews do not persist across steps.
If more than $B_{\mathrm{img}}=24$ images remain, whole messages lose their images, oldest first, until at most 24 remain.

\paragraph{Teacher view.}
The teacher and student use the same memory rules, including the recent window, historical summaries, failure record, observation limits, and image cap.
The teacher receives full tool descriptions, while the student uses concise descriptions.
SFT samples are rendered in the student view.

\begin{figure}[t]
\centering
\fbox{\begin{minipage}{0.94\columnwidth}\scriptsize
\textbf{Summary of an older step.}\\
\texttt{[earlier step: actual\_sequence 7] last action move\_pose, "right arm 486 mm / 0.0° from target, planner failed, not moved", failure\_code planner\_failed. Observation body and images omitted.}

\medskip
\textbf{Failure record.}\\
\texttt{Failed attempts so far (a record kept by the harness, oldest first; not advice):}\\
\texttt{- step 0: grasp\_plan target\_ref=target-x, arm=left refused: stale\_observation (re-ground on the current observation)}\\
\texttt{- step 0: move\_pose right -> (-0.25, 0.08, 0.90) -> right arm 486 mm / 0.0° from target, planner failed, not moved [planner\_failed]}
\end{minipage}}
\caption{Formats of the historical summaries $S_t$ and the failure record $F_t$, with example entries in the format produced by LiteHarness.}
\label{fig:memory-format}
\end{figure}

\subsection{Example of Memory Construction}
\label{sec:app-context-example}

\begin{figure*}[p]
    \centering
    \includegraphics[page=1,width=\textwidth,height=0.93\textheight,keepaspectratio,clip]{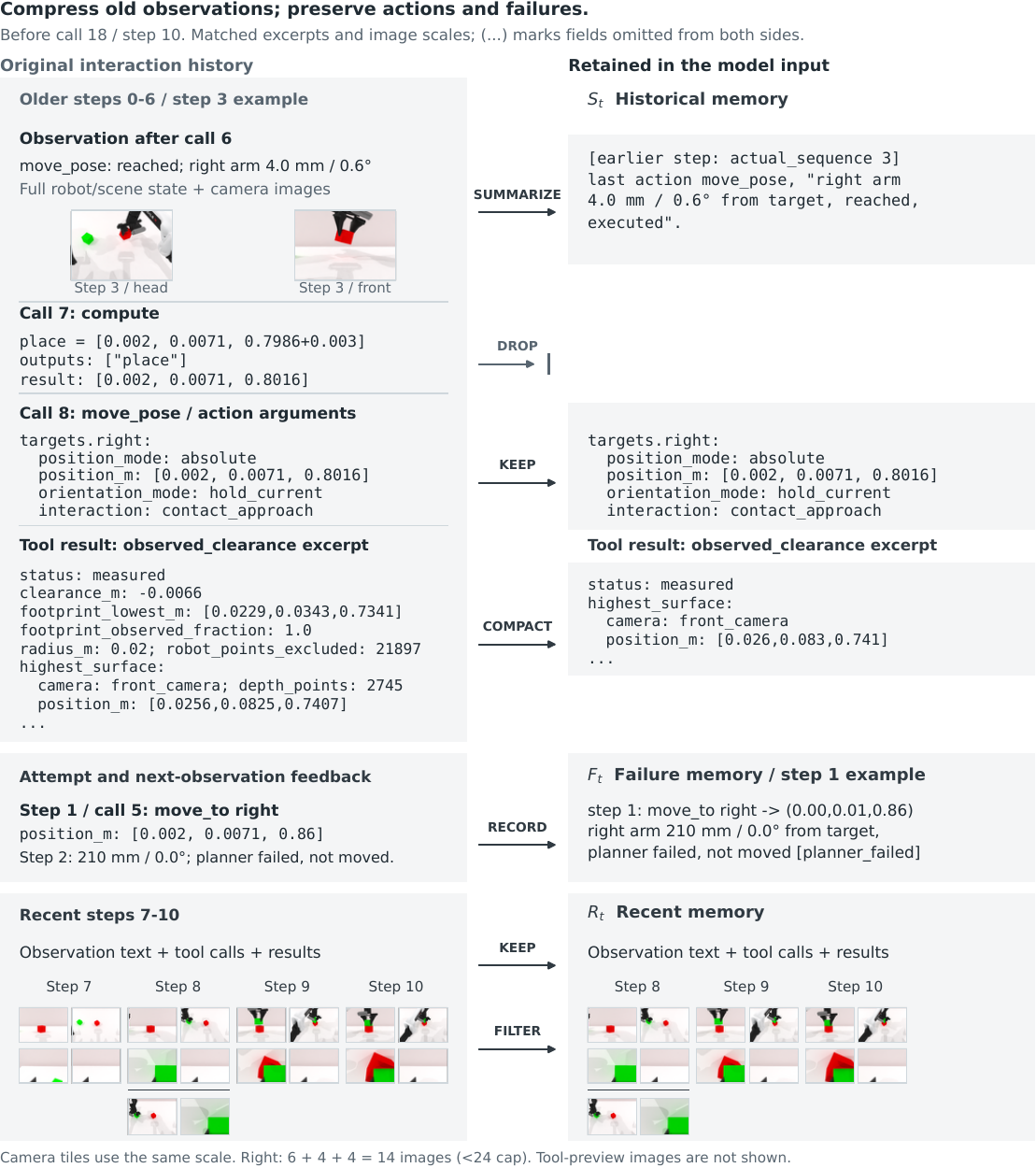}
    \caption{Memory construction before call 18. Arrows map recorded interactions (left) to historical, failure, and recent memory (right). Recent images use matching steps, views, and scales on both sides. Step 8 adds two pre-lift grasp baselines to its four camera views; steps 9 and 10 have four views each. Text excerpts omit fields and abbreviate nesting.}
    \label{fig:memory-construction-overview}
\end{figure*}

Figures~\ref{fig:memory-construction-overview} and~\ref{fig:memory-construction-details} illustrate how LiteHarness constructs a model input from a real interaction history.
We use a \texttt{stack\_blocks\_two} episode of the round-25 RL policy in a clean scene (seed 100011), which succeeds after 20 model calls.
The snapshot is taken immediately before call 18, at observation step 10: the policy has moved the red block to the table center and lowered the green block above it.
The next response opens both grippers; calls 19 and 20 check success and finish the episode.
Step indices identify the observation at which an action is issued, with its physical outcome reported in the next observation.

\newpage
\paragraph{From history to context.}
Figure~\ref{fig:memory-construction-overview} compares recorded interactions with their retained representation.
Historical memory $S_t$ covers steps 0--6.
The top row follows step 3: its observation becomes a short summary of the preceding action's outcome, and the \texttt{compute} query and result are removed.
The subsequent \texttt{move\_pose} call retains the computed placement target, together with a compact tool result.
The middle row traces the step-1 transfer and its step-2 feedback into the separate failure record $F_t$, retaining the attempted target and its outcome; the complete record also contains the attempts at steps 5 and 9.
The bottom row shows the shorter image window within recent memory $R_t$: text spans steps 7--10, while step 8 contributes four camera views plus two grasp-baseline images (head and left wrist, captured after closing and before lifting), and steps 9 and 10 each contribute four camera views.
Step-7 images and tool-preview images are removed; the remaining 14 images are below the 24-image cap.
Failure records can overlap either history window; step 9 also remains in recent memory.

\paragraph{The resulting model input.}
Figure~\ref{fig:memory-construction-details} groups request excerpts into task and retained background, latest interaction and feedback, and the subsequent model response.
Fixed instructions and the task precede $S_t$, the three failure entries in $F_t$, and recent interactions in $R_t$.
The recent excerpt distinguishes the immediate \texttt{submitted} acknowledgement of call 17 from the next observation, which reports that the descent was blocked by contact.
The actual call-18 response opens both grippers and is shown separately, outside the input.
The full request contains 45 messages and 14 images, with 20 tool definitions supplied separately.
JSON excerpts omit fields and abbreviate nesting; four head/front camera crops contrast the scene before the descent with the next observation.

\begin{figure*}[p]
    \centering
    \includegraphics[page=2,width=\textwidth,height=0.93\textheight,keepaspectratio,clip]{figures/memory_construction_example.pdf}
    \caption{Input before call 18 and the next model response. Background excerpts (messages 13--15 and 25) precede the latest interaction (41--44): the descent is submitted, then reported as blocked by contact. The next response opens both grippers. Four recorded camera crops are shown; excerpts omit fields and abbreviate nesting.}
    \label{fig:memory-construction-details}
\end{figure*}

\clearpage

\section{Teacher Demonstrations and Supervised Fine-Tuning}
\label{sec:app-sft}

\subsection{Teacher Data Collection}
\label{sec:app-teacher}

The teacher is GPT-6 Astra, accessed through an OpenAI-compatible API with high reasoning effort and the provider's default sampling.
It acts in the same harness as the student and follows the same memory rules, while receiving full tool descriptions as detailed in Appendix~\ref{sec:app-memory}.
Teacher episodes use ORA's default budgets of 200 model calls and 80 robot decisions.

On RoboTwin, demonstrations are collected in clean scenes on qualified seeds that are disjoint from the evaluation seeds and from the RL seed pool (Appendix~\ref{sec:app-benchmarks}).
On LIBERO-Pro, they are collected on initial states 11--49 of all 80 task cells, adding initial states until a cell has ten successful demonstrations.
Among teacher episodes that ran to completion, 88.7\% succeed on RoboTwin and 88.1\% on LIBERO-Pro; the remaining episodes end with the teacher giving up or exhausting the robot-action budget.
On LIBERO-Pro, success is lower in later batches, which target the cells that remained short of ten demonstrations.

Because the teacher is served through an external API, we monitored each collection batch for changes in its behavior.
For every batch, we tracked the success rate, the mean numbers of model calls and robot decisions per episode, and the change in success between the first and second halves of the batch.
A batch was stopped and inspected when any of these left its reference range, for example when RoboTwin success fell below 70\% or the mean number of model calls exceeded 60 per episode.

\subsection{From Teacher Episodes to SFT Samples}
\label{sec:app-sft-data}

Only episodes that pass the official success check are used.
Successful episodes with more than 80 model calls on RoboTwin, or more than 120 on LIBERO-Pro, are discarded, as such successes are typically reached only after extended unproductive interaction; this removes 11 and 2 episodes, respectively.
On RoboTwin, eight further episodes are held out for validation, leaving the 393 training episodes reported in Sec.~\ref{sec:experimental-setup}.
Each remaining model call yields one training sample.
The input is the context the student would receive at that call, produced by applying the deterministic LiteHarness rendering of Appendix~\ref{sec:app-memory} to the recorded teacher request, and the target is the teacher's response, including its text and tool calls.
Three kinds of calls are excluded as targets: responses rejected by the harness, calls whose tool result was a refusal, and responses containing neither text nor a tool call.
Excluding refused calls prevents the student from imitating invalid arguments, while the teacher's corrected call that follows is kept.
Images are resized to at most 81{,}920 pixels, matching the resolution at which the inference server processes them.
Each sample is rendered with the student's chat template before training to confirm that it matches the input the student receives during evaluation.

Table~\ref{tab:sft-data} summarizes the resulting datasets.
Each training episode contributes one sample per model call except for excluded calls, which account for 91 calls on RoboTwin and 569 on LIBERO-Pro.
RoboTwin samples are longer and contain more images, because two arms and four cameras produce larger observations.
The longest samples reach 37.9K tokens on RoboTwin and 43.7K on LIBERO-Pro, both below the training limit of 49{,}152 tokens.

\begin{table}[h]
    \centering
    \caption{SFT datasets built from successful teacher episodes. Token counts use the Qwen3.5-9B processor and chat template; image tokens follow the processor's resizing. Statistics cover all 401 RoboTwin episodes before the validation split.}
    \label{tab:sft-data}
    \footnotesize
    \setlength{\tabcolsep}{3pt}
    \begin{tabular*}{\columnwidth}{@{\extracolsep{\fill}}lrr@{}}
        \toprule
        & RoboTwin & LIBERO-Pro \\
        \midrule
        Successful episodes in the SFT pool & 412 & 783 \\
        Discarded (too many model calls) & 11 & 2 \\
        Held out for validation & 8 & 0 \\
        Training episodes & 393 & 781 \\
        Training samples & 10{,}306 & 25{,}210 \\
        \midrule
        Model calls / episode (mean, median) & 26.3, 22 & 33.0, 26 \\
        Robot decisions / episode (mean, median) & 10.7, 8 & 16.0, 11 \\
        Prompt tokens / sample (mean) & 22.8K & 18.9K \\
        \quad median, 95th percentile & 23.0K, 29.5K & 18.4K, 26.5K \\
        \quad of which image tokens (mean) & 930 & 425 \\
        Images / sample (mean, max) & 11.7, 24 & 6.6, 20 \\
        \bottomrule
    \end{tabular*}
\end{table}

\subsection{Supervised Fine-Tuning}
\label{sec:app-sft-training}

SFT applies LoRA with rank 16, $\alpha=32$, and dropout 0.05 to all linear projections of the language model, including the attention, MLP, and linear-attention projections of Qwen3.5; the vision encoder is frozen.
The loss is the cross-entropy of the teacher response tokens only, with no loss on the context.
We use AdamW with a constant learning rate of $10^{-4}$, weight decay 0.01, gradient clipping at norm 1.0, and eight samples per update, and train for two epochs.
Samples longer than 49{,}152 tokens are skipped.
Training runs data-parallel on eight NVIDIA H200 GPUs with gradient checkpointing.
The resulting adapter is merged into the base model, which initializes RL and is served for evaluation.

\section{Agentic RL Training}
\label{sec:app-training}

\subsection{Training Procedure}
\label{sec:app-algorithm}

Each RL round alternates between collecting rollouts with the current policy and updating it on the recorded model calls (Algorithm~\ref{alg:round}).
During collection, the policy is served by vLLM behind a recording proxy that stores the token IDs and log-probabilities of every model call, while LiteHarness runs unchanged.
Each model call of an episode becomes one training row, consisting of the context received at that call and the response generated.
Before the update, the trainer rebuilds the prompt of every row from the recorded request with the student's chat template and checks that it matches the recorded token IDs exactly, so the policy is optimized on the same inputs it received during collection.
After the update, the LoRA adapter is merged into the policy, which collects the rollouts of the next round.

\begin{algorithm}[t]
\caption{One RoboHarness-RL training round.}
\label{alg:round}
\small
\begin{algorithmic}[1]
\Require policy $\pi_\theta$, ordered pool of task--initial-state pairs, group size $G$
\State Take the next 12 pairs from the pool
\For{each pair $(x, s_0)$, in parallel}
    \State Run $G$ episodes of $\pi_{\theta_{\mathrm{old}}}$ in LiteHarness from $s_0$ at temperature 1.0; record every model call
    \State Compute each episode's reward $r_i$ (Eq.~\ref{eq:app-reward})
\EndFor
\State Discard groups whose rewards span less than 0.05
\State Compute advantages $\widehat A_{i,t}$ for every recorded call (Eqs.~\ref{eq:group-relative-advantage} and~\ref{eq:app-advantage})
\State Rebuild every prompt and verify it token by token against the recording
\State Update $\theta$ for one epoch over the rows with the loss of Eq.~\ref{eq:multi-turn-grpo}, 512 rows per update
\State Merge the LoRA adapter into $\pi_\theta$
\State Postpone tasks whose group in this round was solved by all $G$ episodes (Appendix~\ref{sec:app-budgets})
\end{algorithmic}
\end{algorithm}

\subsection{Reward and Advantage}
\label{sec:app-reward}

\paragraph{Reward.}
The reward of episode $i$ is its official success, reduced by a bounded penalty for stalling after a failed completion check:
\begin{equation}
    r_i = \mathbb{1}[\text{success}_i] - \lambda\,\frac{\min(e_i, c)}{c},
    \label{eq:app-reward}
\end{equation}
where $e_i$ counts read-only calls made after a negative \texttt{check\_task} result and before the next robot action, beyond the first four of each such run, summed over the episode.
We use $\lambda=0.5$ and $c=160$, so the penalty never exceeds half of the success reward and successful episodes always rank above failed ones within a group.
The penalty targets the most common failure of the SFT policies: after a negative check, the policy keeps querying without acting until the model-call budget is exhausted.
Such episodes frequently form groups in which every rollout fails, which yield no gradient under success alone; the penalty separates failures that keep acting from failures that stall.
Groups whose rewards span less than 0.05 are treated as uninformative and excluded from the update, so a group of failures enters training only when the stall penalty separates its episodes noticeably.

\paragraph{Advantage.}
Advantages are computed within each group of $G=8$ episodes as in Eq.~\ref{eq:group-relative-advantage}, with the baseline taken over episodes rather than over model calls and without normalization by the group's standard deviation.
Every response of an episode receives the episode's advantage, from which the invalid-call penalty is subtracted and to which the episode weight is applied:
\begin{equation}
    \widehat A_{i,t} = \frac{\bar M}{M_i}\left(A_i - p_{i,t}\right),
    \label{eq:app-advantage}
\end{equation}
where $p_{i,t}=0.05$ if call $t$ of episode $i$ is invalid and $p_{i,t}=0$ otherwise.
A call is invalid when it names a tool that does not exist or cannot be parsed; the summed penalty per episode is capped at 0.5.
The weight $\bar M/M_i$ reduces the influence of episodes with more training responses on the token-averaged loss.
Without it, failed episodes, which contain about 60 model calls compared with about 25 for successful ones, would dominate the gradient.
In our implementation, rewards, advantages, and penalties are multiplied by 10; all values in this section and in Table~\ref{tab:hyperparameters} are given before this scaling, with a success reward of 1.

\paragraph{Policy loss.}
The loss in Eq.~\ref{eq:multi-turn-grpo} uses the clipped surrogate of PPO with $\epsilon=0.2$ and dual clipping~\cite{ye2020mastering}, which bounds the loss of tokens with negative advantage at $3\widehat A_{i,t}$ when the probability ratio becomes large.
The loss is averaged over response tokens, and no KL penalty is used.
Rollouts are sampled at temperature 1.0 without top-$p$ or top-$k$ truncation, and training log-probabilities are computed under the same distribution.
Because the inference engine and the trainer compute slightly different probabilities for the same tokens, each token is weighted by the truncated importance ratio
\begin{equation}
    w_{i,t,j} = \min\!\left(\frac{\pi_{\theta_{\mathrm{old}}}(y_{i,t,j}\mid x_i,h_{i,t},y_{i,t,<j})}{\pi_{\mathrm{vLLM}}(y_{i,t,j}\mid x_i,h_{i,t},y_{i,t,<j})},\, 2\right),
\end{equation}
where $\pi_{\mathrm{vLLM}}$ denotes the probability recorded during collection.

\subsection{Hyperparameters}
\label{sec:app-hyperparameters}

Table~\ref{tab:hyperparameters} lists the RL hyperparameters; SFT settings are given in Appendix~\ref{sec:app-sft-training}.

\begin{table*}[t]
    \centering
    \caption{RL hyperparameters.}
    \label{tab:hyperparameters}
    \footnotesize
    \setlength{\tabcolsep}{3pt}
    \begin{tabular*}{0.75\textwidth}{@{\extracolsep{\fill}}lll@{}}
        \toprule
        & RoboTwin & LIBERO-Pro \\
        \midrule
        Initialization & SFT, 393 episodes & SFT, 781 episodes \\
        Reported checkpoint (RL round) & 25 & 25 \\
        Pairs per round / group size $G$ & 12 / 8 & 12 / 8 \\
        Rows per update / epochs per round & 512 / 1 & 512 / 1 \\
        Learning rate & $3\times10^{-5}$ & $3\times10^{-5}$; $5\times10^{-5}$ from round 16 \\
        LoRA rank / $\alpha$ & 16 / 32 & 16 / 32 \\
        LoRA targets & \multicolumn{2}{l}{all linear layers of the language model} \\
        Clipping $\epsilon$ / dual-clip constant & 0.2 / 3.0 & 0.2 / 3.0 \\
        Entropy coefficient $\beta$ & $10^{-3}$ & $10^{-3}$ \\
        KL penalty & none & none \\
        Stall penalty $\lambda$ / cap $c$ / free calls & 0.5 / 160 / 4 & 0.5 / 160 / 4 \\
        Invalid-call penalty / cap & 0.05 / 0.5 & 0.05 / 0.5 \\
        Min.\ reward spread of a trained group & 0.05 & 0.05 \\
        Reward scale & 10 & 10 \\
        Rollout sampling & \multicolumn{2}{l}{temperature 1.0, no truncation} \\
        Rollout correction & \multicolumn{2}{l}{token-level truncated ratio, threshold 2.0} \\
        Max prompt / response tokens & 45{,}056 / 4{,}096 & 53{,}248 / 4{,}096 \\
        Model-call budget $B_{\mathrm{call}}$ & 80--200 per task & 200 \\
        Robot-action budget $B_{\mathrm{act}}$ & 80 & 80 \\
        \midrule
        Evaluation sampling & \multicolumn{2}{l}{temperature 0.7, top-$p$ 0.8, top-$k$ 20} \\
        Hardware & \multicolumn{2}{l}{8 $\times$ NVIDIA H200} \\
        \bottomrule
    \end{tabular*}
\end{table*}

\subsection{Interaction Budgets and Task Scheduling}
\label{sec:app-budgets}

\paragraph{Budgets.}
For each task, the model-call budget $B_{\mathrm{call}}$ is chosen from 80, 120, 160, and 200 calls so that it covers the longest successful teacher and SFT episodes.
This gives $B_{\mathrm{call}}=80$ for most RoboTwin tasks, with the exceptions listed in Table~\ref{tab:budgets}, and $B_{\mathrm{call}}=200$ for all LIBERO-Pro tasks.
The robot-action budget is $B_{\mathrm{act}}=80$ on both benchmarks, and training and evaluation use the same budgets.

\begin{table}[h]
    \centering
    \caption{RoboTwin tasks with a model-call budget $B_{\mathrm{call}}$ above 80.}
    \label{tab:budgets}
    \footnotesize
    \begin{tabular*}{\columnwidth}{@{\extracolsep{\fill}}lr@{}}
        \toprule
        Task & $B_{\mathrm{call}}$ \\
        \midrule
        open\_laptop, blocks\_ranking\_size, blocks\_ranking\_rgb & 120 \\
        put\_bottles\_dustbin, handover\_block, place\_can\_basket, & \\
        \quad place\_mouse\_pad, pick\_diverse\_bottles, pick\_dual\_bottles & 160 \\
        place\_cans\_plasticbox, scan\_object & 200 \\
        \bottomrule
    \end{tabular*}
\end{table}

\paragraph{Scheduling.}
The pool is an ordered list of task--initial-state pairs that cycles through all tasks, and each round takes the next 12 pairs.
On both benchmarks, a task whose most recent group is solved by all eight rollouts is postponed by one cycle, and it is sampled again before it can be postponed a second time.
At a stage boundary, tasks whose rollouts failed in two sampled groups are removed from the pool.
On RoboTwin, this removal takes place at round 15 and leaves 45 of the 50 tasks; postponement is used throughout.

LIBERO-Pro additionally balances task difficulty, because SFT already succeeds on most episodes of the Spatial and Object suites while Goal and Long remain difficult.
Rounds 1--11 cycle through the task cells with postponement.
From round 12, each round samples eight groups from Goal and Long and four from Spatial and Object, after removing six cells whose rollouts failed in two sampled groups.
In all stages, evaluation states are never sampled.

\subsection{Compute}
\label{sec:app-compute}

Each RL round runs on a single node with eight NVIDIA H200 GPUs, alternating between rollout collection and the policy update on the same GPUs.
During collection on RoboTwin, the policy is served by vLLM on three GPUs, one GPU hosts eight instances each of SAM~3 and Contact-GraspNet, and six GPUs run the simulators with eight episodes each, giving 48 parallel episodes; two of the vLLM GPUs also run simulators.
LIBERO-Pro uses two GPUs for vLLM and runs twelve episodes per simulator GPU from round 3, giving 72 parallel episodes.
The update uses verl with FSDP2 on all eight GPUs.
SFT runs data-parallel on the same type of node.

\begin{table}[h]
    \centering
    \caption{Wall-clock time and compute. Round times are means over all rounds; GPU hours count eight GPUs over the wall-clock time of training and exclude evaluation.}
    \label{tab:compute}
    \footnotesize
    \setlength{\tabcolsep}{3pt}
    \begin{tabular*}{\columnwidth}{@{\extracolsep{\fill}}lrr@{}}
        \toprule
        & RoboTwin & LIBERO-Pro \\
        \midrule
        Rollout collection / round (min) & 39.6 & 30.2 \\
        Policy update / round (min) & 84.7 & 94.3 \\
        Total / round (min) & 124.4 & 124.5 \\
        Episode time, success / failure (min) & 4.9 / 13.5 & 4.2 / 15.5 \\
        \midrule
        RL: rounds / wall clock (h) & 25 / 51.8 & 25 / 51.9 \\
        RL GPU hours & 415 & 415 \\
        SFT: wall clock (h) / samples per second & 4.9 / 1.18 & 10.5 / 1.33 \\
        SFT GPU hours & 39 & 84 \\
        \bottomrule
    \end{tabular*}
\end{table}

Table~\ref{tab:compute} summarizes the cost.
A round takes about two hours on both benchmarks, of which rollout collection accounts for 32\% on RoboTwin and 24\% on LIBERO-Pro.
The update dominates because every model call of a long episode is a separate training row with its own multimodal prompt, averaging 27.5K tokens on RoboTwin and 23.7K on LIBERO-Pro.
Failed episodes take three to four times as long as successful ones, as most of them run until a budget is exhausted.
The teacher is accessed through an API, so teacher collection uses local GPUs only for simulation and perception.

\begin{table}[h]
    \centering
    \caption{Cost of one policy update on the same three RoboTwin task groups (24 rollouts per setting) under hierarchical memory and ORA's compressed full history. GPU-hours measure scoring and gradient-update computation, while wall-clock times also include startup and reconstruction overhead.}
    \label{tab:memory-cost}
    \footnotesize
    \setlength{\tabcolsep}{3pt}
    \begin{tabular*}{\columnwidth}{@{\extracolsep{\fill}}lrrr@{}}
        \toprule
        & Hierarchical & Full history & Ratio \\
        \midrule
        Rows (dropped as too long) & 1{,}227 (0) & 981 (98) & -- \\
        Tokens in the update & 32.3M & 48.4M & 1.50$\times$ \\
        Mean prompt tokens per row & 26K & 56K & 2.1$\times$ \\
        Update compute (GPU hours) & 2.43 & 4.00 & 1.65$\times$ \\
        Update wall clock (min) & 33 & 51 & 1.55$\times$ \\
        \bottomrule
    \end{tabular*}
\end{table}

\paragraph{Training cost of memory.}
To measure how memory affects training cost, we take the round-25 RoboTwin policy and the next three task groups in its schedule (lift\_pot, shake\_bottle, and stack\_blocks\_two; eight rollouts each).
For hierarchical memory, we use the 24 rollouts recorded during training; for ORA's compressed full history, we sample 24 new rollouts from the same policy on the same groups.
We then run one complete update, scoring and gradient step, on the same eight-GPU node and record tokens and time per batch; the updated weights are discarded.
The full-history rows require raising the prompt limit from 45{,}056 to 77{,}824 tokens, and 10\% of them still exceed it (Table~\ref{tab:memory-cost}).
Over the prompts of the 250 clean evaluation episodes, 76\% of full-history calls exceed 45{,}056 tokens, whereas no hierarchical row is dropped in the update.
Per million tokens, the longer sequences cost only 3--12\% more, so most of the difference comes from the 1.5$\times$ as many tokens entering the update.
Since the update accounts for more than two thirds of each round, a complete round would cost about 1.45$\times$ as much under the compressed full history.
This is a single measurement on RoboTwin used only for timing; it does not compare the success of policies trained under each memory.

\subsection{Training Dynamics}
\label{sec:app-dynamics}

\begin{figure*}[t]
    \centering
    \includegraphics[width=\textwidth]{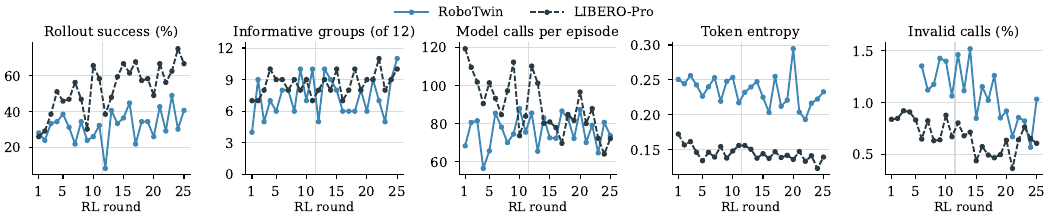}
    \caption{RL training dynamics per round, computed from the training rollouts (96 episodes per round) and the update logs. The vertical line marks the LIBERO-Pro scheduling stage starting at round 12.}
    \label{fig:training-dynamics}
\end{figure*}

Figure~\ref{fig:training-dynamics} shows the training statistics of every round.
Rollout success is measured on the training pairs of each round, which change with the schedule, so it fluctuates more than evaluation success.
On LIBERO-Pro, it rises from 31\% over the first three rounds to 65\% over the last five, although the later stages concentrate on difficult cells; on RoboTwin, it rises from 28\% to 38\% over the same spans.
Each round yields 7.4 informative groups out of 12 on RoboTwin and 8.6 on LIBERO-Pro on average, and every round contains at least four.
Episodes become shorter on LIBERO-Pro, from 110 to 75 model calls on average, consistent with the lower execution overhead in Sec.~\ref{sec:behavioral-analysis}, while the invalid-call rate stays below 1.6\% on both benchmarks.

The optimization stays close to the rollout policy.
Token entropy remains between 0.19 and 0.30 on RoboTwin and between 0.12 and 0.17 on LIBERO-Pro without a downward trend, so the policy does not collapse onto a narrow set of responses.
The approximate KL divergence between consecutive policies stays below $2\times10^{-3}$ per round, the fraction of clipped tokens below 0.5\%, and the mean importance weight of the rollout correction within $10^{-4}$ of one.
Each round performs 4 to 16 optimizer updates, depending on the number of rows in informative groups.

\section{Qualitative Examples}
\label{sec:app-qualitative}

This section complements Fig.~\ref{fig:feedback-execution} with an RL episode on RoboTwin that recovers from repeated failures, four paired SFT and RL episodes on LIBERO-Pro, and an analysis of where RL episodes still fail.
The RoboTwin example includes a call-level table; the paired LIBERO-Pro cases align key events and subsequent actions across policies.
The examples illustrate the statistics in Sec.~\ref{sec:behavioral-analysis} and are not representative samples.

\subsection{Recovery in a RoboTwin Episode}
\label{sec:app-qual-episode}

\begin{figure*}[t]
    \centering
    \includegraphics[width=\textwidth,height=0.50\textheight,keepaspectratio,clip]{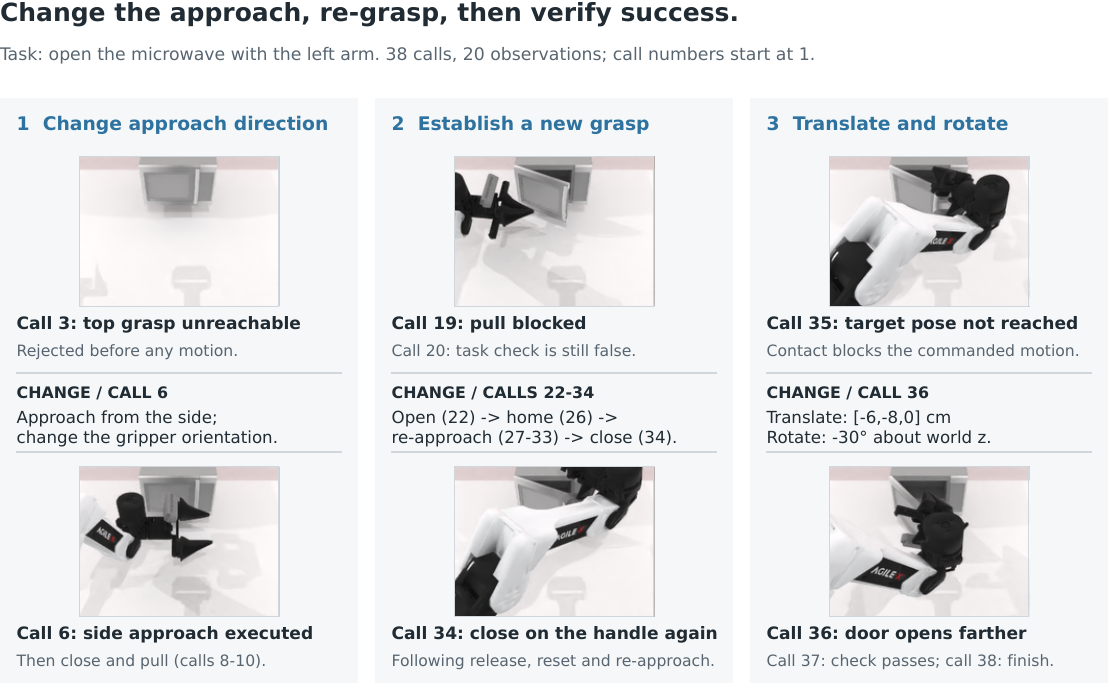}
    \caption{Recovery in one \texttt{open\_microwave} episode (Table~\ref{tab:microwave-episode}). Each column pairs a failed attempt with the subsequent adjustment: change approach direction, release/reset/re-grasp, or translate and rotate. The middle strip specifies the action and call numbers. All images are recorded head-camera crops after the labelled calls; the rejected grasp leaves the initial scene unchanged.}
    \label{fig:microwave-frames}
\end{figure*}

Table~\ref{tab:microwave-episode} traces the \texttt{open\_microwave} episode of Fig.~\ref{fig:framework} (round-25 policy, seed 100018, clean scene), whose instruction is ``Pull the medium-sized microwave with angled walls open with the left arm.''
The episode contains 38 model calls over 20 observation steps, each response containing exactly one tool call.
Figure~\ref{fig:microwave-frames} highlights three concrete changes, with call numbers starting at 1.
First, call 3 rejects the top-grasp candidate before motion; call 6 changes the gripper orientation and approaches the handle from the side, followed by closing and short outward pulls in calls 8--10.
Second, after further pulls remain blocked, including call 19, the policy opens the gripper in call 22, returns the arm home in call 26, re-approaches through calls 27--33, and closes on the handle again in call 34.
Third, call 35 is still blocked; call 36 commands a world-frame translation of $[-0.06,-0.08,0]$\,m together with a $-30^\circ$ rotation about the world $z$ axis, continuing the opening motion.
Call 37 checks success and call 38 finishes the episode.

Seven of the 38 responses include a short statement before the tool call, reproduced verbatim:
\begin{itemize}\footnotesize
\item[2:] I'll use the left gripper to grasp the vertical handle, then pull gently and follow the door's swing. I'll check that the door moves with the gripper before continuing to a wide-open position.
\item[6:] The proposed grasp was rejected before any motion. I'll try a side approach to the handle instead, keeping the jaws across its width and pulling outward.
\item[8:] The approach stopped short with the gripper visibly touching the door, so I won't push farther. I'll close at this position, then make a short outward pull to test whether the handle follows.
\item[11:] The door moved outward with the gripper, and a gap is now visible. I'll follow a curved path around the microwave's left edge, pulling farther while keeping the handle between the jaws.
\item[19:] The door has swung open, but the gripper stopped short of the requested pose. I'll change the pull direction to follow the remaining arc around the visible hinge, rather than repeat the blocked straight-line pull.
\item[21:] The door is still only partially open. I'll measure the exposed upper hinge area, then pull along its arc using a fresh grasp rather than pushing farther with the current grip.
\item[36:] The door is visibly farther open, and the gripper is now positioned around the exposed handle. I'll continue along the same arc toward a wide-open position.
\end{itemize}
These statements are brief plans stated together with the tool call; thinking is disabled, and no response contains step-by-step reasoning.

\begin{table}[t]
    \centering
    \caption{Model calls of the \texttt{open\_microwave} episode, grouped by observation step. Outcomes of world-changing calls appear in the next observation.}
    \label{tab:microwave-episode}
    \scriptsize
    \setlength{\tabcolsep}{2.5pt}
    \begin{tabular*}{\columnwidth}{@{\extracolsep{\fill}}rlp{0.62\columnwidth}@{}}
        \toprule
        Step & Calls & Tools and outcomes \\
        \midrule
        0 & 1--6 & \texttt{vision\_ground}, \texttt{grasp\_plan}, \texttt{grasp\_execute} (refused: pre-grasp unreachable), \texttt{mark\_point}, \texttt{compute}, \texttt{move\_pose} (side approach) \\
        1 & 7 & \texttt{move\_pose} toward the handle; stops 76\,mm short, blocked by contact \\
        2 & 8 & \texttt{set\_gripper}: close on the handle \\
        3--4 & 9--10 & \texttt{move\_pose}: short outward pulls; the door opens a gap \\
        5 & 11 & \texttt{move\_pose} along a curved path; blocked \\
        6 & 12--15 & \texttt{vision\_ground}, \texttt{mark\_point}, \texttt{observe}, \texttt{move\_pose}; blocked \\
        7 & 16 & \texttt{move\_pose}; blocked \\
        8 & 17--19 & \texttt{check\_task} (false), \texttt{mark\_point}, \texttt{move\_pose} with a new pull direction; blocked \\
        9 & 20--22 & \texttt{check\_task} (false), \texttt{mark\_point} on the hinge area, \texttt{set\_gripper} \\
        10 & 23--25 & \texttt{vision\_ground}, \texttt{grasp\_plan}, \texttt{move\_pose} \\
        11 & 26 & \texttt{recover\_pose} \\
        12--15 & 27--31 & \texttt{move\_pose} (four times), with \texttt{vision\_ground} at step 14 \\
        16 & 32--34 & \texttt{mark\_point}, \texttt{compute}, \texttt{set\_gripper} \\
        17--18 & 35--36 & \texttt{move\_pose} (twice), continuing along the arc \\
        19 & 37--38 & \texttt{check\_task} (true), \texttt{finish\_episode} \\
        \bottomrule
    \end{tabular*}
\end{table}

\subsection{Paired Episodes on LIBERO-Pro}
\label{sec:app-qual-pairs}

\begin{figure*}[p]
    \centering
    \includegraphics[width=\textwidth,height=0.43\textheight,keepaspectratio,clip]{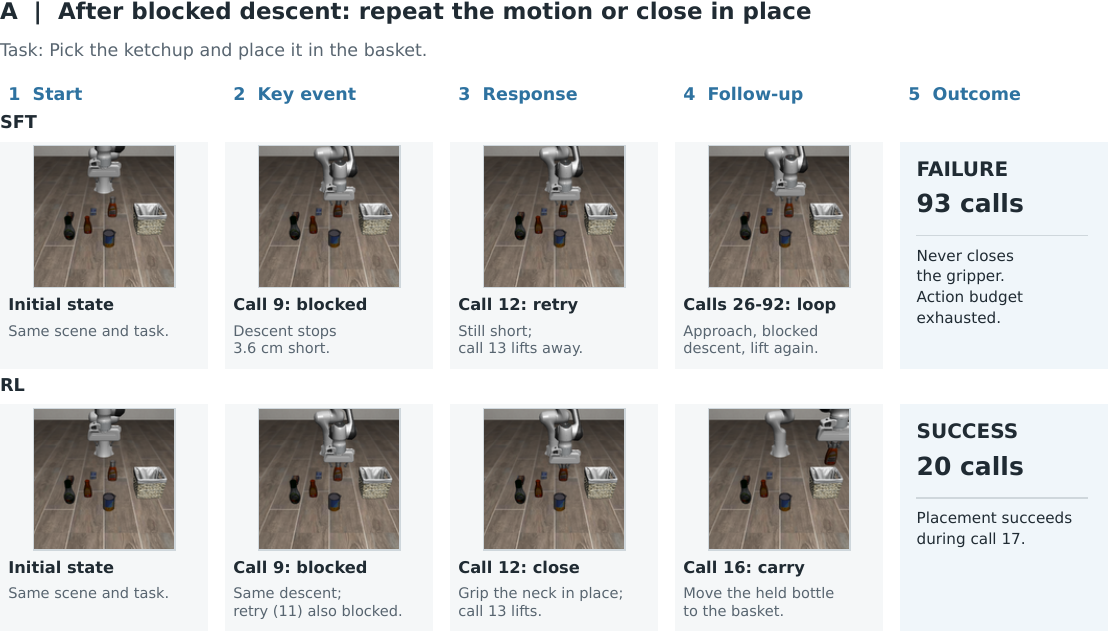}
    \caption{Case A. Object, Task variant, task 3, initial state 10. After the same blocked descent, SFT repeats approach/lift motions without closing the gripper, while RL closes in place and completes placement. Columns align event stages; call indices start at 0.}
    \label{fig:case-a}
\end{figure*}

\begin{figure*}[p]
    \centering
    \includegraphics[width=\textwidth,height=0.43\textheight,keepaspectratio,clip]{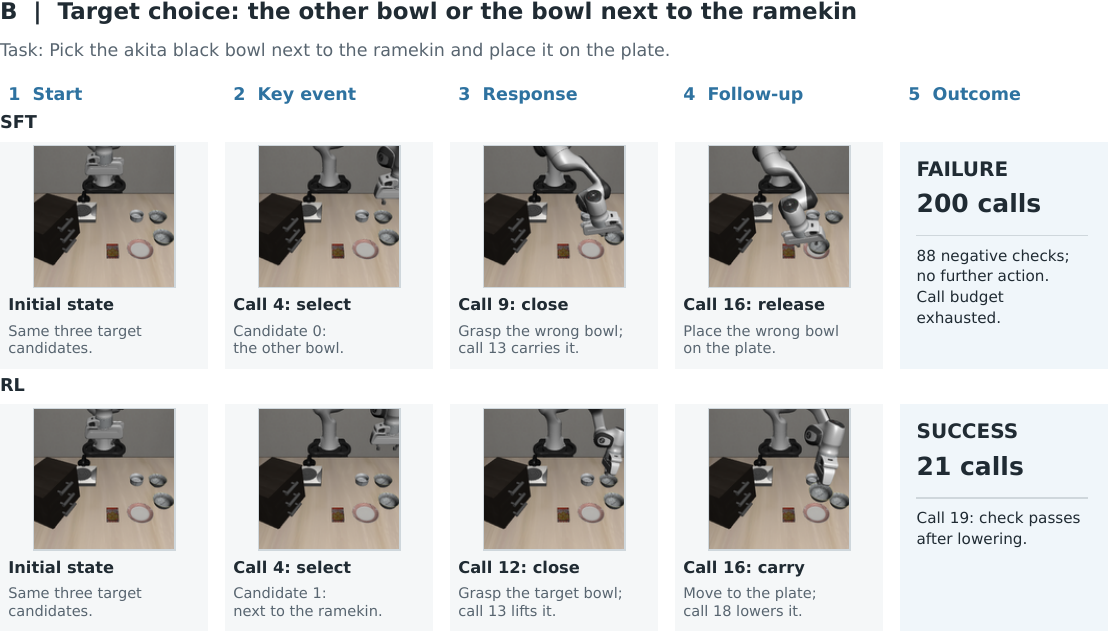}
    \caption{Case B. Spatial, Task variant, task 8, initial state 9. The policies choose different bowls from the same candidate set. SFT places the wrong bowl and repeats checks; RL selects the bowl next to the ramekin and succeeds. Columns align event stages; call indices start at 0.}
    \label{fig:case-b}
\end{figure*}

\begin{figure*}[p]
    \centering
    \includegraphics[width=\textwidth,height=0.43\textheight,keepaspectratio,clip]{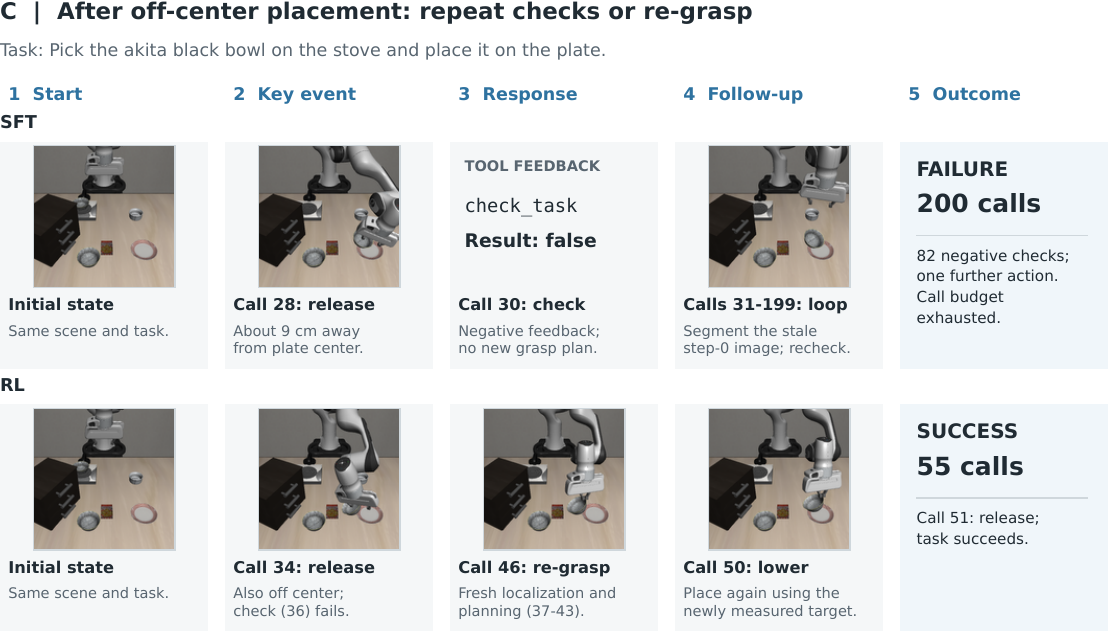}
    \caption{Case C. Spatial, Task variant, task 6, initial state 1. Both first placements are off center. SFT mostly repeats checks using an old observation; RL re-localizes, re-grasps, and places again successfully. Columns align event stages; call indices start at 0.}
    \label{fig:case-c}
\end{figure*}

\begin{figure*}[p]
    \centering
    \includegraphics[width=\textwidth,height=0.43\textheight,keepaspectratio,clip]{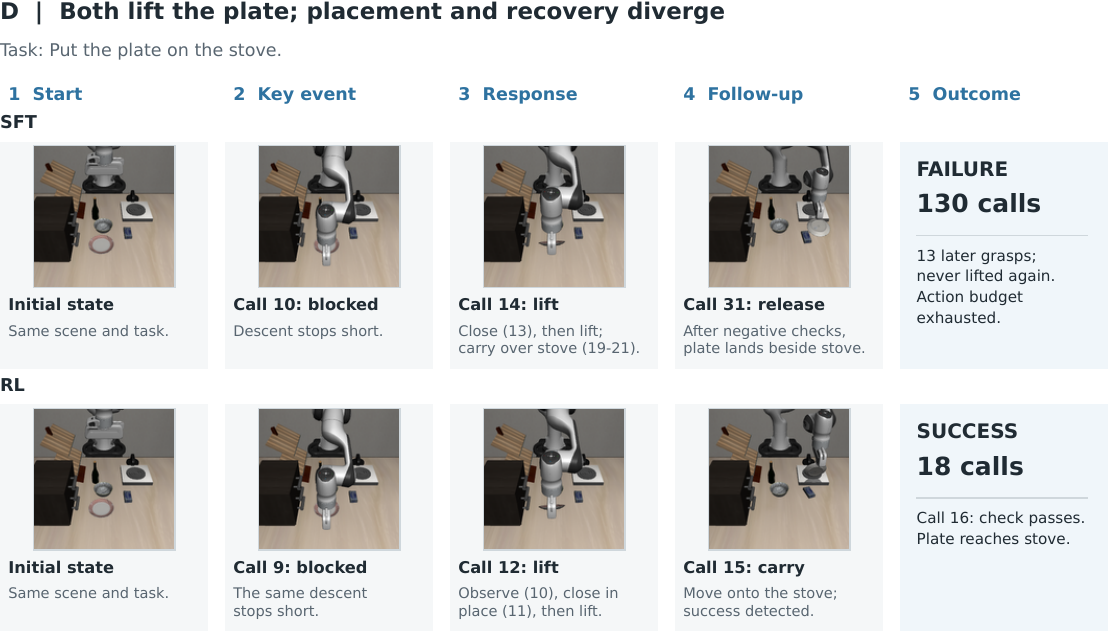}
    \caption{Case D. Goal, Task variant, task 1, initial state 10. Both policies lift the plate. SFT releases it beside the stove and fails to lift it again; RL carries it onto the stove and passes the success check. Columns align event stages; call indices start at 0.}
    \label{fig:case-d}
\end{figure*}

Figures~\ref{fig:case-a}--\ref{fig:case-d} compare SFT and RL episodes from the same task and initial state; call indices start at 0, and the captions quote the task instructions.
Cases A--C were selected from 52 pairs read in full, spanning episodes where only RL succeeds, where only SFT succeeds, and where both fail.
Case D was selected from pairs where only RL succeeds because the two policies issue nearly identical calls up to the same negative feedback, so the episodes differ only in the response to it.
These cases illustrate differences between SFT and RL in target selection and responses to execution feedback.
Each figure aligns the starting scene, a key event, the response, subsequent follow-up actions, and the episode outcome; the aligned events need not occur at the same call index.

\subsection{Where RL Episodes Still Fail}
\label{sec:app-qual-failures}

On LIBERO-Pro, we categorize failed episodes by the furthest stage reached (Table~\ref{tab:failure-stages}), counting a grasp as held when the gripper does not fully close after closing on an object.
After RL, nearly all failures reach the placement stage: in 88\% of failed episodes the policy has grasped and placed an object, yet the official check fails, compared with 77\% for SFT.
Episodes that never close the gripper or never place a held object nearly disappear (58 versus 8), whereas closing without holding an object remains similar (20 versus 18).
On the 183 initial states that both policies fail, RL performs more pick-and-place attempts (3.2 versus 2.4 per episode, $p=8\times10^{-8}$) and tries more grasp locations (2.9 versus 2.2, $p=3\times10^{-7}$), so RL failures result from attempts that do not reach the required placement rather than from stopping early.
Neither policy declares success prematurely: no LIBERO-Pro episode ends with a success claim that the official check rejects.

\begin{table}[h]
    \centering
    \caption{Furthest stage reached in failed LIBERO-Pro episodes.}
    \label{tab:failure-stages}
    \footnotesize
    \setlength{\tabcolsep}{3pt}
    \begin{tabular*}{\columnwidth}{@{\extracolsep{\fill}}lrr@{}}
        \toprule
        & SFT & RL \\
        \midrule
        Failed episodes & 338 & 214 \\
        Never closed the gripper & 20 & 2 \\
        Closed but never held an object & 20 & 18 \\
        Held but never placed & 38 & 6 \\
        Placed, yet the check fails & 260 (77\%) & 188 (88\%) \\
        \bottomrule
    \end{tabular*}
\end{table}

\section{Limitations}
\label{sec:app-limitations}

\paragraph{Simulation.}
All experiments are conducted in simulation.
Agentic RL collects many multi-call rollouts in every round, which simulation provides at scale but would be difficult to collect on real robots; evaluating transfer to real robots is left to future work.

\paragraph{Dependence on the teacher.}
We initialize RL from teacher demonstrations because the base model rarely succeeds on its own: the untrained Qwen3.5-9B reaches 20.0\% on clean RoboTwin and 1.8\% on LIBERO-Pro, which leaves too few successful rollouts for RL to learn from.
RL therefore inherits the coverage of the demonstrations to some extent.

\paragraph{Remaining failure modes.}
RL improves how the policy uses the tools, while perception, grasp proposals, and motion planning come from fixed external models.
Most remaining LIBERO-Pro failures place the object near, but not at, the required pose (Appendix~\ref{sec:app-qual-failures}), and long-horizon tasks and randomized bimanual scenes show the largest remaining gaps (Secs.~\ref{sec:main-results} and~\ref{sec:comparison}).

\paragraph{Statistical scope.}
Each RL run of 25 rounds takes about 415 GPU hours (Table~\ref{tab:compute}), and evaluating one checkpoint takes 250 or 800 multi-call episodes.
We therefore train one RL run per benchmark and evaluate each reported checkpoint once; the paired tests in Appendix~\ref{sec:app-significance} capture variation across episodes, not across training seeds.

\end{document}